%% file: ieee.tex
\documentclass[final]{IEEEtran}
\usepackage{amssymb}
\usepackage{graphicx}
\usepackage{mathrsfs,amsmath}
\usepackage{tikz}
\usepackage[english]{babel}
\usepackage[utf8]{inputenc}
\usepackage{algorithm}
\usepackage[noend]{algpseudocode}
\usepackage{multirow}
\usepackage{graphicx}
\usepackage{url}
\usepackage{xcolor}
\usepackage{soul}
\usepackage{filecontents,lipsum}
\usepackage{dblfloatfix}
\usepackage[utf8]{inputenc}
\usepackage{dtk-logos}

\usepackage{subfig}  
\usepackage[compact]{titlesec}         
\usepackage[bottom]{footmisc} 
\AtBeginDocument{
  \setlength\abovedisplayskip{3pt}
  \setlength\belowdisplayskip{3pt}
  \setlength{\textfloatsep}{3pt}}

\usepackage[compact]{titlesec}
\titlespacing{\section}{0pt}{*1}{*0}
\titlespacing{\subsection}{0pt}{*1}{*0}
\titlespacing{\subsubsection}{0pt}{*0}{*0}
\usepackage{breqn}

\usepackage{amsmath}
\usepackage{amsfonts}
\usepackage{tabularx}
\usepackage[noadjust]{cite}

\usepackage{comment}

\ifCLASSINFOpdf
\else
\fi

\usepackage{placeins}
\makeatletter
\renewcommand{\fnum@figure}{Fig. \thefigure}
\makeatother
\addto\captionsenglish{}                     
\usepackage{etoolbox}
\makeatletter
\patchcmd{\@maketitle}
  {\addvspace{0.5\baselineskip}\egroup}
  {\addvspace{-1.5\baselineskip}\egroup}
  {}
  {}
\makeatother

\usepackage{enumitem}

\usepackage[compact]{titlesec}         
\usepackage[bottom]{footmisc} 
\AtBeginDocument{
  \setlength\abovedisplayskip{2pt}
  \setlength\belowdisplayskip{2pt}
  \setlength{\textfloatsep}{2pt}}

\usepackage[compact]{titlesec}
\titlespacing{\section}{0pt}{*1}{*0}
\titlespacing{\subsection}{0pt}{*1}{*0}
\titlespacing{\subsubsection}{0pt}{*0}{*0}
\usepackage{breqn}

\begin{document}

\newcommand{\mycaption}[1]{\stepcounter{figure}\raisebox{0pt}{\footnotesize Fig. \thefigure.\hspace{1pt} #1}}
%

\title{A Fully Spiking Hybrid Neural Network for Energy-Efficient Object Detection}
%
%
%

\author{Biswadeep~Chakraborty,
        Xueyuan~She,~\IEEEmembership{Student Members,~IEEE,}
        Saibal~Mukhopadhyay,~\IEEEmembership{Fellow,~IEEE}
\thanks{
This material is based on research supported by the Defense Advanced Research Projects Agency (DARPA) under contract number HR0011-17-2-0045. The views and conclusions contained herein are those of the authors and should not be interpreted as necessarily representing the official policies or endorsements, either expressed or implied, of DARPA.

B.Chakraborty, X.She and   S.   Mukhopadhyay   are   with   the   Department  of  Electrical  and  Computer  Engineering,  Georgia  Institute  of  Technology,   Atlanta,   GA,   30332   USA   e-mail:(biswadeep@gatech.edu; xshe@gatech.edu; saibal.mukhopadhyay@ece.gatech.edu)}
}

\maketitle

\begin{abstract}
This paper proposes a Fully Spiking Hybrid Neural Network (FSHNN) for energy-efficient and robust object detection in resource-constrained platforms. The network architecture is based on a Spiking Convolutional Neural Network using leaky-integrate-fire neuron models. The model combines unsupervised Spike Time-Dependent Plasticity (STDP) learning with back-propagation (STBP) learning methods and also uses Monte Carlo Dropout to get an estimate of the uncertainty error. FSHNN provides better accuracy compared to DNN based object detectors while being more energy-efficient. It also outperforms these object detectors, when subjected to noisy input data and less labeled training data with a lower uncertainty error.
\end{abstract}


%
\IEEEpeerreviewmaketitle

\input{tex/intro}

\input{tex/related}

\input{tex/motivation}

\input{tex/system}

\input{tex/perf_eval}

\input{tex/conclusions}
\bibliographystyle{IEEEtran}
\bibliography{egbib.bib}

\input{tex/supple}
\end{document}

%% file: tex/intro.tex
\section{Introduction}
Spiking Neural Networks (SNNs) are widely acclaimed as the third generation of neural networks given their low power consumption and close similarity in emulating how the brain works. 
There are two main aspects of spiking neural networks that make them attractive. Firstly, in SNNs, neurons communicate with each other through isolated, discrete electrical signals called spikes, as opposed to continuous signals, and work in continuous-time instead of discrete-time \cite{maass1997networks}. This spike-based inference methodology makes the spiking neural network energy-efficient\cite{panda2020toward,stone2018principles,merolla2014million, carrillo2012advancing}. In addition to this, spiking neural networks provides a novel unsupervised learning methodology using Spike Time Dependent Plasticity (STDP)\cite{bliss1993synaptic}. STDP is a bio-plausible unsupervised learning mechanism that instantaneously manipulates the synaptic weights based on the temporal correlations between pre-and post-synaptic spike timings. STDP can also help improve robustness to noise in the inputs and assists in learning with less labeled data \cite{tavanaei2016bio}, \cite{panda2018learning}. 
These properties are attractive for real-world computer vision applications where the input can be imperfect and the labeled training data can be sparse.

The recent advances in Deep Neural Network-to-Spiking Neural Network (DNN-to-SNN) conversion mechanisms have enabled design of SNN with performance comparable to DNN in large datasets like ImageNet \cite{deng2009imagenet}.
However, despite the recent developments, most of the existing works on spiking networks are primarily limited to classification. Many real-time autonomous applications such as lightweight drones and edge robots, where energy-efficient inference is crucial, have object detection as the primary task. Hence, an SNN for object detection will facilitate deployment of SNNs in such applications. 
Recent work on spiking based object detection \cite{kim2020spiking} primarily uses the backpropagated SNN model and is trained on small networks (Tiny YOLO). This leads to a comparatively low accuracy compared to standard object detection networks. Moreover, the network does not use STDP learning limiting noise robustness or learning from less amount of labeled data. 


\begin{figure}
    \centering
    \includegraphics[width = \linewidth]{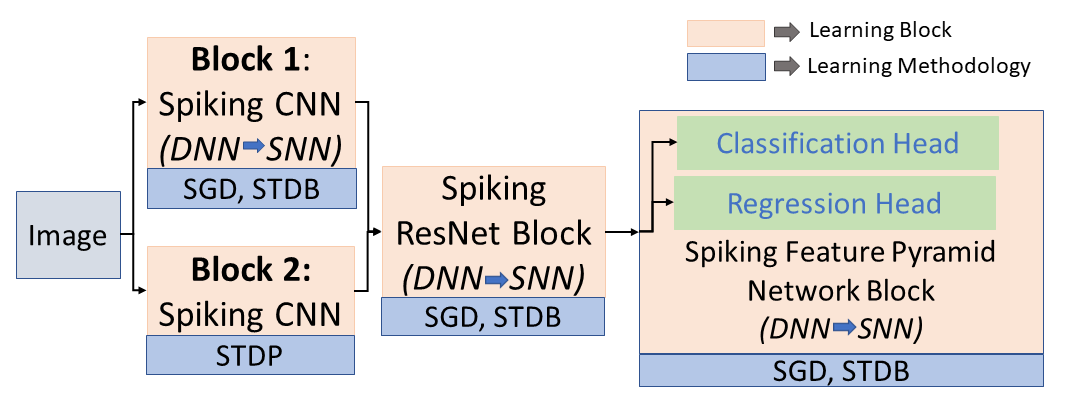}
    \caption{General Block Diagram for Different Architectures}
    \label{fig:block1}
\end{figure}

\par In this paper, we propose a novel fully spiking neural network-based object detector using hybridization of unsupervised STDP based learning and supervised backpropagated learning. 
The proposed FSHNN network uses the baseline architecture of an object detection DNN, such as RetinaNet, as the backbone. The low-level feature space of the classifier is enhanced by fusing an auxiliary CNN block and a set of Spiking Convolution Layers pre-trained using STDP. Finally, we use a DNN-to-SNN conversion technique \cite{rathi2020enabling} while keeping the STDP-trained layer fixed to get the final object detector.  The paper makes the following key contributions:
\begin{itemize}
    \item We develop a fully spiking hybrid neural network for object detection integrating Stochastic Gradient Descent (SGD) and STDP based learning.
    \item We show that the STDP-based learning results in a more generalizable SNN model than a SGD-trained model.  
    \item We adapt the Monte Carlo dropout methodology \cite{nessler2013bayesian} to quantify uncertainty for FSHNN and use that to discriminate true positives from false-positive.
    \item We show that the hybridization of SGD-based and STDP-based learning within a SNN improves accuracy, improves generalizability, reduces uncertainty, and increases energy efficiency.

\end{itemize}



\par We evaluate the performance of FSHNN for object detection on MSCOCO dataset. Our experiments show that the FSHNN with six layers of STDP achieves an mAP of 0.426 compared to 0.388 achieved by the standard RetinaNet. We also show that FSHNN outperforms RetinaNet when tested with noisy input images and trained with limited labeled data. Using the Monte Carlo Dropout method, we further demonstrate that FSHNN based object detector has a lower uncertainty compared to RetinaNet. Also, we illustrate that the FSHNN based object detector is more generalizable compared to the standard DNN based RetinaNet.
Moreover, as the network uses spiking operations during inference, it naturally enhances the energy-efficiency of object detection.


\par The rest of the paper is structured as follows: Section II discusses the Related Works associated with this paper; Section III revolves around the architectures and baselines used in the evaluation; Section IV deals with the Experiments undertaken and the results obtained thereby; Section V summarizes the observations and discusses the conclusions we arrived at from the experiments.

%% file: tex/related.tex
\section{Related Works}
\subsection{Object Detection:}

Object detection has become a critical problem of many computer vision tasks as it requires the estimation of both the category to which an object belongs and its spatial location. Single-stage detection  models like Single-shot  multi-box  detector (SSD) \cite{liu2016ssd} and YOLO \cite{redmon2018yolov3} achieve fast inference with the cost of lower mean Average Precision (mAP) and mean Average Recall (mAR). RetinaNet \cite{lin2017focal} overcomes these problems as it improves on the Faster-RCNN framework with the help of a focal loss and using ResNet+ Feature Pyramid Network (FPN) as the backbone network to extract features. 

\subsection{Spiking Object Detection: }Though object detection has been an important field of interest in computer vision, there is not much work for the use of spiking neural networks in object detection. Recently, Kim et al. implemented a spiking version of the YOLO object detection architecture \cite{kim2020spiking}. They used channel-wise normalization and signed neurons with imbalanced thresholds to provide fast and accurate information transmission for deep SNNs. However, the proposed spiking YOLO object detector had a much lower performance than the standard DNN based YOLO and RetinaNet models.

\subsection{DNN-to-SNN conversion:}
Recent works have proposed weight normalization and threshold balancing methods to obtain minimal loss for accuracy during the conversion process. In this paper, we use the threshold balancing method \cite{sengupta2019going} that gives near-lossless conversion performance for deep architectures like ResNet on complex Imagenet dataset.

Diehl et al. used the relationship between dynamics of the spiking neuron and the ReLu activation function for a fast DNN-to-SNN conversion \cite{diehl2015fast}, achieving a nearly lossless conversion on the MNIST dataset using parameter normalization.

 Rueckauer et al. proposed the SNNToolbox framework \cite{rueckauer2016theory} which improved the SNN scores after conversion. 
Miquel et al. used the algorithm and implementations suggested by Rueckauer et al. for a DNN-to-SNN conversion of the RetinaNet  \cite{miquel2021retinanet}.
 In this paper, we use the Spiking-RetinaNet implementation done by Miquel et al. as the initialization step. We use a rate-based encoding scheme for the DNN-to-SNN conversion process i.e. the neurons generate spike-trains whose rate approximates the analog activation corresponding to the original DNN. This encoding becomes more accurate as the simulation duration is increased entailing a higher resolution at the price of scaling up the computational cost.

\subsection{STDP based Unsupervised learning in SNN}
Spike Timing Dependent Plasticity (STDP) \cite{huang2014associative} is a variant of the Hebbian unsupervised learning algorithm. STDP describes the changes of a synaptic weight according to the relative timing of pre and post-synaptic spikes. In STDP, if a presynaptic spike precedes a post-synaptic one, a synaptic weight is potentiated.

There are various spiking neural networks (SNNs) to solve object recognition and image classification tasks by converting the traditional DNNs into spiking networks by replacing each DNN computing unit with a spiking neuron \cite{diehl2016conversion}. Kheradpisheh et al. proposed a STDP based spiking deep convolutional neural network for object recognition \cite{kheradpisheh2018stdp}. Wu et al. similarly proposed a spatio-temporal backpropagation for training Spiking Neural Networks \cite{wu2018spatio}. This type of spiking neural networks primarily aim to reduce the energy consumption in DNNs while achieving a similar accuracy.

\subsection{Back-propagation based Supervised Learning in SNN} 

 Backpropagation-based supervised learning for spiking neural networks has played a critical role in improving their performance. 
Rathi et. al used a hybrid training technique that combines ANN-SNN conversion and spike-based BP that reduces the latency and helps in better convergence \cite{rathi2020enabling}.  The authors used the ANN-SNN conversion as an initialization step followed by spike-based backpropagation (STDB) incremental training. The authors also demonstrated the approach of taking a converted SNN and incrementally training it using backpropagation. This hybrid approach improves the energy-efficiency and accuracy compared to standard models trained with either only spike-based backpropagation or only converted weights.

\subsection{Generalizability of Neural Networks}

Recent works \cite{jacot2018neural, huang2020deep, chen2020generalized} have analyzed overparameterized (wide) neural networks from a theoretical perspective by connecting them to reproducing kernel Hilbert spaces. The papers have also shown that under proper conditions, the weights of a well-trained overparameterized network remain very close to their initialization. Thus, during training the model searches within some class of reproducing kernel functions, where the associated kernel is called the “neural tangent kernel” which only depends on the initialization of the weights. 
\par Another line of research has worked on comparing the generalization error while using different training strategies \cite{csimcsekli2020hausdorff}. The authors show that we can control the generalization error of a training algorithm using the Hausdorff dimension of its trajectories. 
Gurbuzbalaban et al. showed that depending on the structure of the Hessian of the loss at the minimum, and the choices of the algorithm parameters $\eta, b$, the SGD iterates will converge to a heavy-tailed stationary distribution \cite{gurbuzbalaban2020heavy}.

Recent works have worked on determining the generalization characteristics of the object proposal generation that is the first step in detection models \cite{wang2020leads}. 
A more generalizable object proposal can help to scale detection models to a larger number of classes with fewer annotations. Thus, the paper studies how a detection model trained on a small set of source classes can provide proposals that generalize to unseen classes.

 

%% file: tex/motivation.tex
\section{Motivation}

In this section, we discuss the primary motivations for using a hybrid spiking neural network-based object detector. We discuss the generalization advantages we can obtain by fusing features learned using STDP with the SGD process.

\subsection{LIF and STDP Dynamics}

Spiking neural network uses biologically plausible neuron
and synapse models that can exploit temporal relationship between spiking events \cite{moreno2015causal},\cite{lansdell2018spiking}. There are different models that
are developed to capture the firing pattern of real biological
neurons. We choose to use Leaky Integrate Fire (LIF) model
in this work described by:
\begin{align}
\tau_m \frac{dv}{dt} &= -v(t) +R I(t) \\
v&=v_{\text {reset }}, \text { if } v>v_{\text {threshold }}
\end{align}

where, $v$, the membrane potential, and $\tau_m$, the membrane time constant of the neuron. R is the resistance of the integrate-and-fire model and $I(t)$ is the sum of the current signal from all synapses that connect to the neuron.
In SNN, two neurons connected by one synapse are referred
to as pre-synaptic neurons and post-synaptic neurons. Conductance of the synapse determines how strongly two neurons are connected and learning is achieved through modulating
the conductance using STDP \cite{bliss1973long}, \cite{gerstner1993spikes}. With two operations of
STDP, namely, long-term potentiation (LTP) and long-term depression
(LTD), SNN can extract the causality between spikes
of two connected neurons from their temporal relationship.
More specifically, LTP is triggered when post-synaptic neuron
spikes closely after a pre-synaptic neuron spike, indicating a causal relationship between the two events. On the other hand,
when a post-synaptic neuron spikes before pre-synaptic spike
arrives or without receiving a pre-synaptic spike at all, the
synapse goes through LTD. 



In a STDP model, the change in synaptic weight induced by the pre-and post-synaptic spikes at times $t_{pre}, t_{post}$ are defined by:

\begin{align}
		\Delta W_{ij}=
		\begin{cases}
			a_{LTP} \left(W_{ij}-W_{LB}\right) \left(W_{UP}-W_{ij}\right) & \quad t_j - t_i \leq 0,\\
			a_{LTD} \left(W_{ij}-W_{LB}\right) \left(W_{UP}-W_{ij}\right) & \quad  t_j - t_i > 0,\\
		\end{cases}
	\end{align}
	where $i$ and $j$ refer to the post- and pre-synaptic neurons, respectively,
	$\Delta w_{ij}$ is the amount of weight change for the synapse connecting the two neurons,
	and $a_{LTP}$, and $a_{LTD}$ scale the magnitude of weight change. Besides,
	$\left(W_{ij}-W_{LB}\right)\times \left(W_{UP}-W_{ij}\right)$ is a stabilizer term which
	slowes down the weight change when the synaptic weight is close to the weight's lower ($W_{LB})$
	and upper $(W_{UB})$ bounds.




\subsection{Transfer Learning and Feature Fusion}

 In this subsection, we motivate the feature fusion of STDP and SGD processes for object detection. We argue that features learned using STDP can complement the standard SGD making the training simpler, more generalizable, and robust. Zeiler et. al. observed that the lower layers of a DNN converged much faster compared to the higher layers \cite{zeiler2014visualizing}. They also showed that small transformations in the input image impact the lower layers more than the higher layers. Donahue et al. have empirically demonstrated that a convolutional network trained on a large dataset can be successfully generalized to other tasks where less data is available \cite{donahue2014decaf}. These observations motivated us to use additional layers in the initial stages of the backbone classifier to capture the low-level features. 
Yosinski et al. showed that lower level features can be transferred to different networks to get superior performance \cite{yosinski2014transferable}. Developing on these ideas, we utilize the spiking convolutional layers trained using the unsupervised STDP method since it can better apprehend local features compared to the standard DNN. 

Thus, we show the generalizability of STDP compared to SGD that would subsequently motivate us to demonstrate the generalization of spiking neural network models.

\subsection{Generalizability of STDP}

In this subsection, we compare the generalization bounds of a spiking convolutional neural network to its DNN counterpart with the same architecture. Jacot et al. showed that the generalizability of an infinitely wide neural network is dependent on the initialization where a better initialization improves the generalizability of the network \cite{jacot2018neural}. 

Simsekli et al. demonstrated that the generalization error is controlled by the uniform Hausdorff dimension of the learning algorithm, with the constants inherited from the regularity conditions  \cite{csimcsekli2020hausdorff}. The authors further showed that the Hausdorff dimension is controlled by the tail behavior of the process, with heavier tails implying less generalization error. Therefore, we study the tail indices of the trajectories of the STDP and SGD learning methodologies as a measure of its Hausdorff Dimension. 

Recent literature has shown that the STDP based learning in SNN follows an Ornstein–Uhlenbeck (OU) process \cite{leen2012approximating}. On the other hand, SGD follows a Feller process \cite{simsekli2020hausdorff}. So, we compare the tail indices of the two stochastic processes to study their heavy-tailed natures. 

We use stochastic differential equations (SDEs) to formulate local plasticity rules for parameters $\theta_{i}$ that control synaptic connections (if $\left.\theta_{i}>0\right)$ and synaptic weights $w_{i}=\exp \left(\theta_{i}-\theta_{0}\right)$
\begin{align}
d \theta_{i}=\left(b \frac{\partial}{\partial \theta_{i}} \log p^{*}(\boldsymbol{\theta})\right) d t+\sqrt{2 T b} \cdot d \mathcal{W}_{i} \nonumber \\ \text { where } d \mathcal{W}_{i} \text { denotes an infinitesimal }
\end{align}
step of a random walk (Wiener process), drift diffusion $\mathrm{b}=$ learning rate, $\mathrm{T}=$ temperature (T=1 until last slide)

The Fokker-Planck (FP) equation tracks the resulting evolution of the SNN configurations $\boldsymbol{\theta}$ over time, yielding the stationary distribution $\frac{1}{z}p^*(\boldsymbol{\theta})^{1/T}$ which follows the Ornstein–Uhlenbeck process \cite{leen2012approximating}
\begin{align}
\frac{\partial}{\partial t} p_{F P}(\boldsymbol{\theta}, t)=\sum_{i}-\frac{\partial}{\partial \theta_{i}}\left(\left(b \frac{\partial}{\partial \theta_{i}} \log p^{*}(\boldsymbol{\theta})\right) p_{F P}(\boldsymbol{\theta}, t)\right) \nonumber \\ + \frac{\partial^{2}}{\partial \theta_{i}^{2}}\left(T b p_{F P}(\boldsymbol{\theta}, t)\right)
\end{align}

The practically relevant forms of the target distribution $p^*(\boldsymbol{\theta})$ of network configurations depend on the type of learning. For this case, we consider the unsupervised learning method described in 
\cite{kappel2015synaptic}.
We perform a tail index analysis of STDP in deep SNN, similar to the method adopted in \cite{simsekli2019tail}.

Several works have explored estimation of the tail-index of an extreme-value distribution \cite{hill1975simple, pickands1975statistical}. However, most of these methods fail to estimate the tail index of $\alpha$-stable distributions \cite{mittnik1996tail, paulauskas2011once}. We consider the estimation method proposed by Mohammadi et. al. \cite{mohammadi2015estimating} which has a faster convergence rate and smaller asymptotic variance than the aforementioned methods. The basic approach is discussed below.

In the following, $f_{X}(.)$ and $F_{X}(.)$ denote the probability density function and distribution function of a random variable $X,$ respectively, and $\xi_{p}^{X}$ satisfies$p=P\left(X \leq \xi_{p}^{X}\right)$.

We know for the one-dimensional case, if $X_{1}, \ldots, X_{n}$ be a sequence of i.i.d. random variables with probability density function $f_{X}(.),$ and $X_{1: n} \leq \cdots \leq X_{n: n}, \frac{j}{n} \rightarrow p \in(0,1),$ then
\begin{equation}
\lim_{n \rightarrow+\infty} \sqrt{n}\left(X_{j: n}-\xi_{p}^{X}\right) \rightarrow_{D} N\left(0, \frac{p(1-p)}{\left(f_{X}\left(\xi_{p}^{X}\right)\right)^{2}}\right)
\end{equation}
where $" \rightarrow_{D} "$ denotes convergence in distribution.

We consider $|X|_{1: n} \leq \cdots \leq|X|_{n: n}$ such that $|\boldsymbol{x}|=\left(\sum_{k=1}^{\lambda} x_{k}^{2}\right)^{1 / 2} \forall x=\left(x_{1}, \ldots, x_{d}\right) \in \mathbb{R}^{d} .$ Consider an i.i.d. sequence of strictly $\alpha$ -stable random vectors $\boldsymbol{X}_{i}, i=1,2, \ldots, m$ with dimension $d \geq 1 .$ Let $\boldsymbol{Y}=\sum_{i=1}^{m} \boldsymbol{X}_{i} .$ 
As Mohammadi et. al. has shown \cite{mohammadi2015estimating}, 
$$
\left(\xi_{p}^{\log |Y|}-\xi_{p}^{\log \left|X_{1}\right|}\right) / \log m=1 / \alpha
$$
These relations were used to construct an estimator for $1 / \alpha $ as follows:

 Let $\left\{X_{i}\right\}_{i=1}^{K}$ be a collection of random variables with $X_{i} \sim \mathcal{S} \alpha \mathcal{S}(\sigma)$ and $K=K_{1} \times K_{2} .$ Define $Y_{i} \triangleq \sum_{j=1}^{K_{1}} X_{j+(i-1) K_{1}}$ for $i \in\left[1, K_{2}\right] .$ Then, the estimator
\begin{align}
\widehat{\frac{1}{\alpha}} \triangleq \frac{1}{\log K_{1}}\left(\frac{1}{K_{2}} \sum_{i=1}^{K_{2}} \log \left|Y_{i}\right|-\frac{1}{K} \sum_{i=1}^{K} \log \left|X_{i}\right|\right)
\label{eq:estimator}
\end{align}
converges to $1 / \alpha$ almost surely, as $K_{2} \rightarrow \infty$.

In order to estimate the tail-index $\alpha$ at iteration $k,$ we first partition the set of data points $\mathcal{D} \triangleq\{1, \ldots, n\}$ into many disjoint sets $\Omega_{k}^{i} \subset \mathcal{D}$ of size $b,$ such that the union of these subsets give all the data points. Formally, for all $i, j=1, \ldots, n / b,\left|\Omega_{k}^{i}\right|=b, \cup_{i} \Omega_{k}^{i}=\mathcal{D},$ and $\Omega_{k}^{i} \cap \Omega_{k}^{j}=\emptyset$ for $i \neq j$
We then compute the full gradient $\nabla f\left(\mathbf{w}_{k}\right)$ and the stochastic gradients $\nabla \tilde{f}_{\Omega_{k}^{i}}\left(\mathbf{w}_{k}\right)$ for each minibatch $\Omega_{k}^{i} .$ Finally. we compute the stochastic gradient noises $U_{k}^{i}\left(\mathbf{w}_{k}\right)=\nabla \tilde{f}_{\Omega_{i}}\left(\mathbf{w}_{k}\right)-\nabla f\left(\mathbf{w}_{k}\right),$ vectorize each $U_{k}^{i}\left(\mathbf{w}_{k}\right)$ and concatenate
them to obtain a single vector, and compute the reciprocal of the estimator given in Eq. \ref{eq:estimator}. We use $K=p n / b$ and $K_{1}$ is the divisor of $K$ that is the closest to $\sqrt{K}$ for our experiments.

\begin{table}[]
\centering
\caption{Table showing the hyperparameters used for the LIF and the STDP models}
\label{tab:stdpparams}

\begin{tabular}{|c|c||c|c|}
\hline
\textbf{\begin{tabular}[c]{@{}c@{}}STDP\\ Parameters\end{tabular}}                                                 & \textbf{Value} & \textbf{\begin{tabular}[c]{@{}c@{}}LIF\\ Parameters\end{tabular}}                                                                           & \textbf{Value} \\ \hline
\textit{\begin{tabular}[c]{@{}c@{}}LTP Learning \\ Rate $(a_{LTP})$ \end{tabular} }                                              &  0.004             &  $R$                                                        &       1      \\ \hline
\textit{\begin{tabular}[c]{@{}c@{}}LTD Learning \\ Rate $(a_{LTD})$ \end{tabular} }                                               &  0.003      & $\tau_m$                                                    &      10             \\ \hline
$W_{LB}$         &      0 & $\mathbf{v_{\text{threshold}}}$                                                 &       -40             \\ \hline
$W_{UP}$         &      1 & $\mathbf{v_{\text{reset}}}$                                                 &        -80            \\ \hline
\end{tabular}
\end{table}

\begin{figure}
    \centering
    \includegraphics[width = \columnwidth]{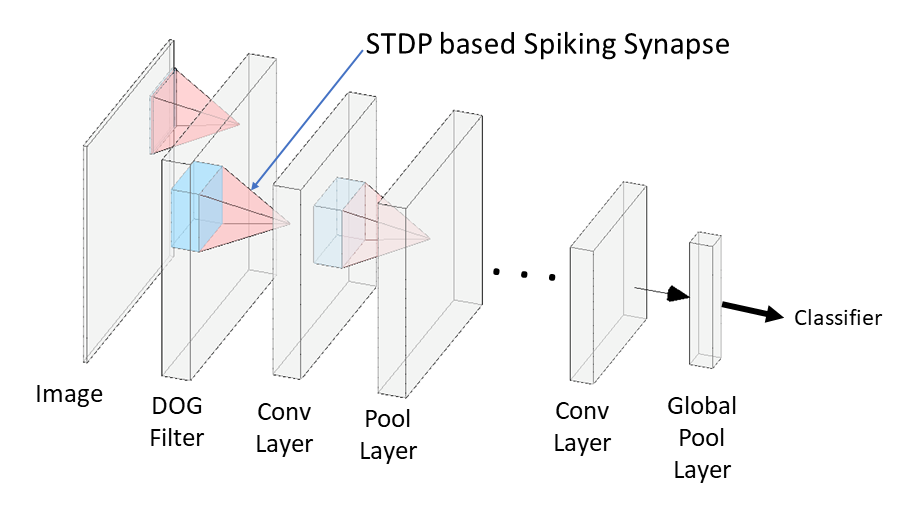}
    \caption{Architecture used for tail index analysis}
    \label{fig:tail_arch}
\end{figure}

Similar to the analysis done by Simsekli et. al, we evaluate the tail-index of the standard CNN trained with the SGD and the spiking convolutional neural networks trained with STDP processes \cite{simsekli2019tail}. The generalized architecture model used for this experiments are shown in Fig.  \ref{fig:tail_arch} . For the spiking convolution layers, we use STDP based learning method as described in Fig. \ref{fig:tail_arch} similar to the training methodology proposed by  Kheradpisheh et al. \cite{kheradpisheh2018stdp}\cite{she2020safe}. The parameters used for the simulation are enlisted in Table \ref{tab:stdpparams}. It is to be noted that this training is completely unsupervised until the final global pooling and classifier which are trained with the labels are given. We used a deep SNN, comprising several convolutional, trained using STDP, and pooling layers. For the SGD based DNN approach, we use the same architecture as described in Fig. \ref{fig:tail_arch}, but used standard Convolutional Neural Networks (CNNs) instead and trained the network using SGD on a cross-entropy loss function. The SGD was trained with a batch size of 32, a learning rate of 0.004, and a momentum of 0.9. Both the networks were trained for 300 epochs before their tail indices were estimated. The experiment is repeated with varying numbers of layers and is evaluated on the MNIST dataset and the tail indices are reported in Table \ref{tab:tail}. We observe that the STDP process has a smaller value of $\alpha$ for each of the cases thus implying that the distribution has a heavier tail and thus is more generalizable.

\begin{table}[]
\centering
\caption{Comparison of Tail Indices $\alpha$ of SGD, STDP for varying depth in Convolutional Networks\cite{simsekli2019tail}}
\label{tab:tail}
\resizebox{0.6\columnwidth}{!}{%
\begin{tabular}{|c|c|c|}
\hline
\textbf{\begin{tabular}[c]{@{}c@{}}No. of\\ Conv. Layers\end{tabular}} & \textbf{SGD} & \textbf{STDP} \\ \hline
\textbf{3} & 1.297 & 1.206 \\ \hline
\textbf{4} & 1.291 & 1.193 \\ \hline
\textbf{5} & 1.284 & 1.188 \\ \hline
\textbf{6} & 1.273 & 1.181 \\ \hline
\end{tabular}%
}
\end{table}


%% file: tex/system.tex
\section{Proposed Architecture}

In this section, we introduce the novel Fully Spiking Hybrid Neural Network-based Object Detector. Since fusion improves the learning ability of low-level features and STDP is more generalizable than SGD, we introduce a model that fuses the STDP and the SGD based learning methods.

\subsection{Image to Spike Conversion and Output Decoding}

\begin{figure}
    \centering
    \includegraphics[width = \columnwidth]{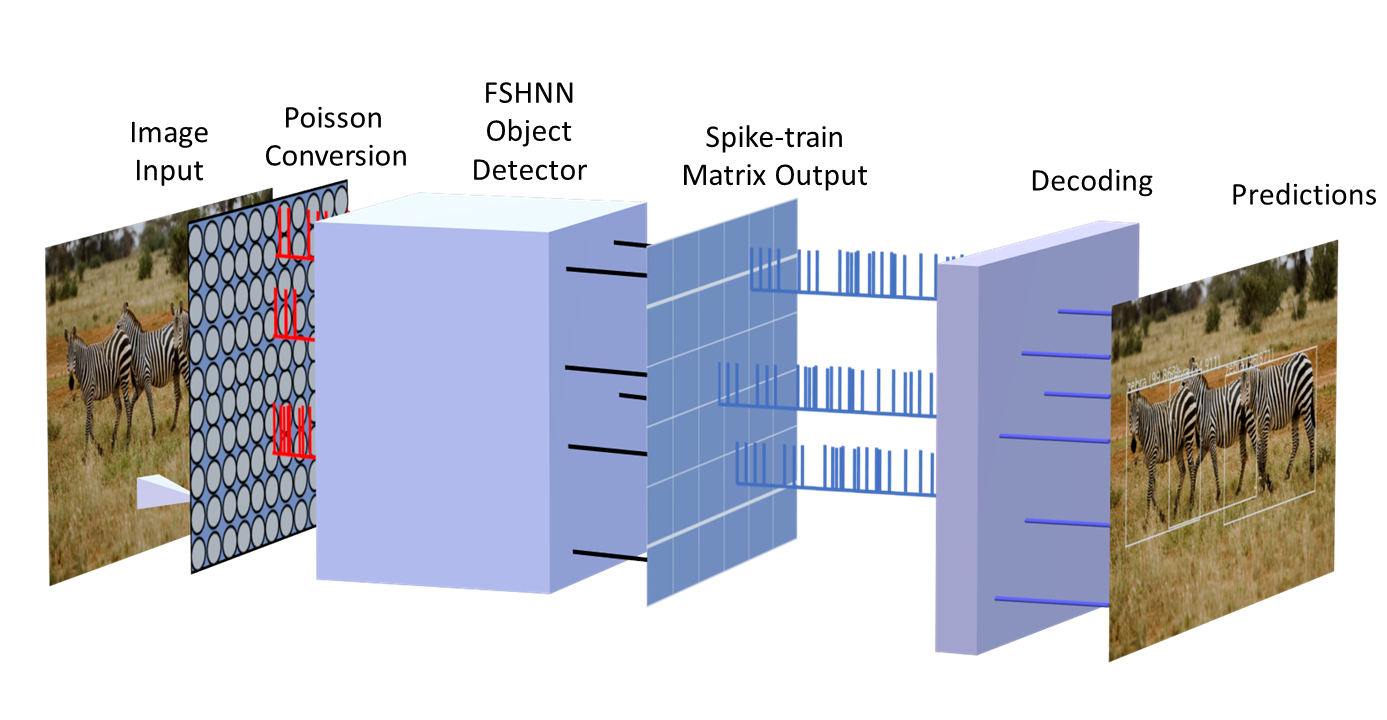}
    \caption{Illustrative scheme of the FSHNN simulation for a specific time window}
    \label{fig:enc_dec}
\end{figure}

SNNs process Poisson rate-coded input spike trains, wherein, each pixel in an image is converted to a Poisson-distribution-based spike train with the spiking frequency proportional to the pixel value.
As shown in Fig 3, the input layer consists of a single neuron per image pixel. Each input is a Poisson spike-train, which is fed to the excitatory neurons of the second layer. The rates of spikes of each neuron are proportional to the intensity of the corresponding pixel. The intensity values of the image are converted to Poisson-spike with firing rates proportional to the intensity of the corresponding pixel. The obtained spiking object detector is
exposed to the input image within a given time window where the image is fed to the model as a matrix of constant currents as shown in Fig. \ref{fig:enc_dec}. 
It then propagates through the network driving the final
layer to a steady spiking rate. After these rate values are
denormalized they can be processed and decoded.

\textbf{Decoding: } The decoding is performed in the same way as the original ANN output activations, in post-processing, by selecting the output neuron with the largest number of spikes. In order to facilitate the success of this decoding rule, in the training phase, the postsynaptic neuron corresponding to the correct label is assigned a desired output spike train, while a zero output is assigned to the other postsynaptic neurons. 

However for the object detection problem, the regression problem to determine the bounding boxes poses a challenge as it requires high numerical precision. Low firing rate in neurons and lack of an efficient implementation method of leaky-ReLU in SNNs often leads to a performance degradation for such high-precision tasks as shown by Kim et al. \cite{kim2020spiking}. Hence, as shown by the authors, we used a fine-grained channel-wise normalization to enable fast and efficient information transmission in deep SNNs. This normalizes the weights by the maximum possible activation in a channel-wise manner instead of the conventional layer-wise manner. Channel-norm normalizes extremely small activations such that the neurons can transmit information accurately in a short period of time. These small activations are critical in regression problems and significantly affect the model’s accuracy. The output decoding is thus done by accumulating the $V_{mem}$ to decode the spike trains from the regression and classification heads



\begin{figure}
    \centering
\includegraphics[width = 0.7\columnwidth]{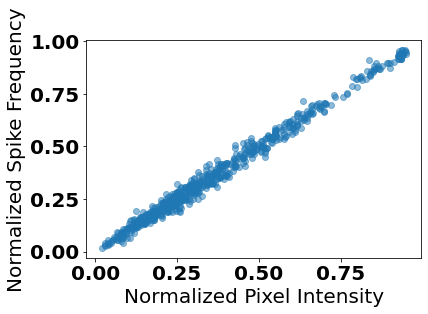}
         \caption{Plot for the Normalized Spike Frequency vs the Normalized Pixel Intensity}
         \label{fig:spike_freq}
\end{figure}

\begin{figure*}
    \centering
    \includegraphics[width = \linewidth]{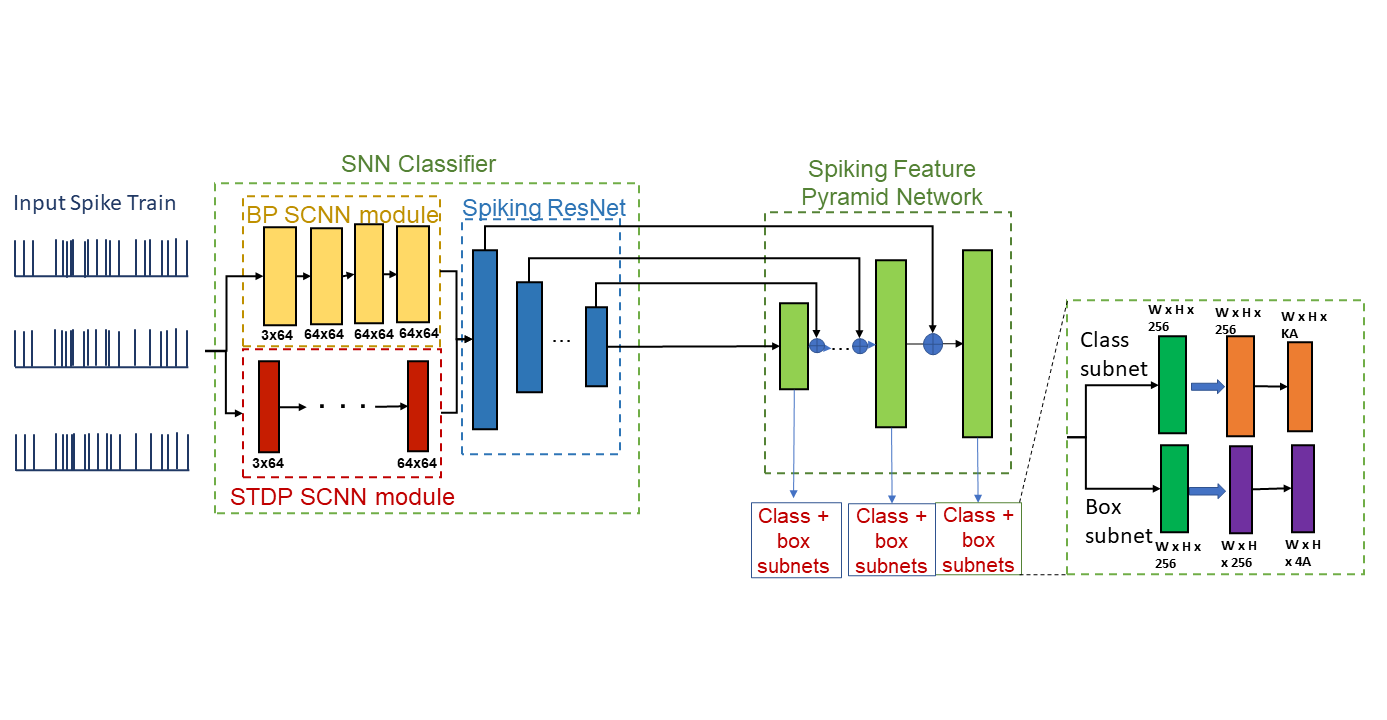}
    \caption{Architectural block diagram showing the different parts of the FSHNN spike based object detector with varying STDP layers}
    \label{fig:block2}
\end{figure*}

\begin{figure}
    \centering
    \includegraphics[width = \linewidth]{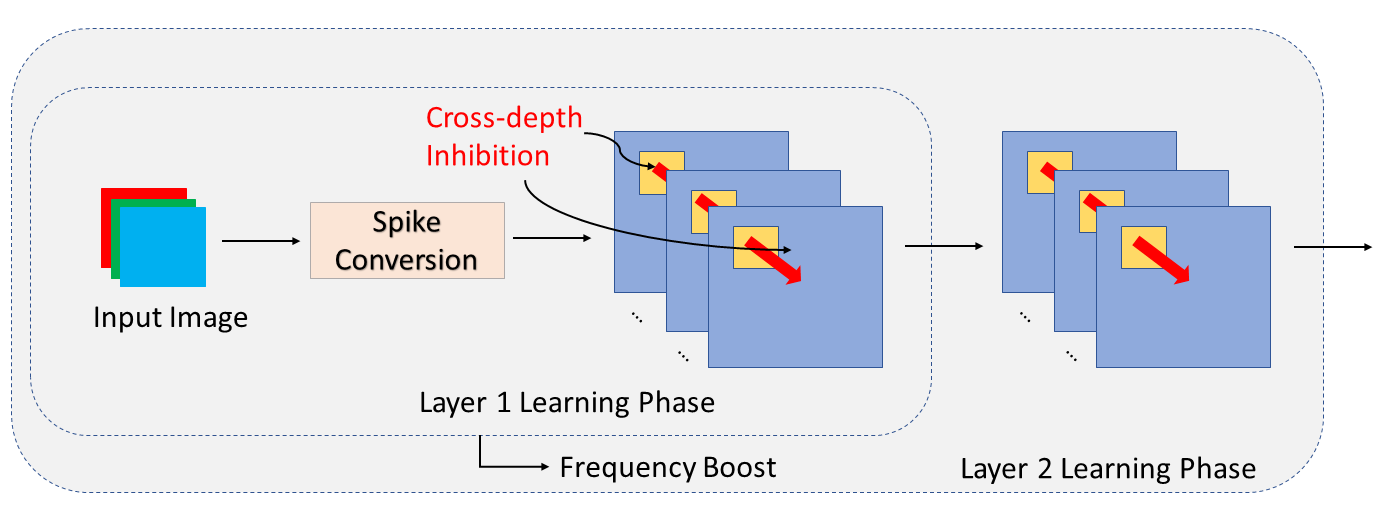}
    \caption{The architecture of the spiking convolutional module for feature extraction and layer-by-layer learning process}
    \label{fig:cross}
\end{figure}

\subsection{Fully Spiking Hybrid Object Detector Architecture}

For the FSHNN based Object Detector model, we take the DNN based RetinaNet model\cite{lin2017focal} with a backbone network of ResNet 101. We augment the ResNet layers with an Auxilliary CNN block and an STDP block as shown in Fig. \ref{fig:block1}. We keep the STDP block frozen and convert the rest of the DNN to a Spiking network using a DNN to SNN conversion method as described in \cite{rathi2020enabling}. Hence, we get a fully spiking object detector with augmented STDP layers and an auxiliary CNN block in the backbone. The motivation for such a model is to compare the benefits of adding an unsupervised training methodology in the object detection network.
  
For this paper, we use a ResNet-101 as the backbone network. For \textbf{Block 1} in FSHNN, we use four convolutional layers converted from their DNN counterparts. 
\textbf{Block 2} for FSHNN is composed of the STDP layers. As we repeat the experiments with multiple STDP layers, we keep on increasing a $64 \times 64$ filter in each case at the end of the last Spiking Convolutional Layer in \textbf{Block 2} as shown in Fig \ref{fig:block1}.  The detailed filter sizes for the \textbf{Blocks 1,2} and the ResNet are given in Fig \ref{fig:block2}. However, it is to be noted that  the \textbf{STDP SCNN module} shown in Fig.  \ref{fig:block2} is a general representation for the STDP block with varying Spiking Convolutional layers trained using STDP. We perform experiments by varying the number of Spiking Convolutional layers from 3 to 6.

\subsection{STDP based Spiking CNN architecture} 

The architecture of the unsupervised STDP based convolutional module is shown in Fig. \ref{fig:cross}. The architecture differs from the conventional DNN based architectures in the following respects. 

The connections are made with plastic synapses following the STDP learning rule. When the neuron in the SCNN layer spikes, an inhibitory signal is sent to the neurons at the same spatial coordinates across all depths in the same layer. This inhibition facilitates the neurons at the same location to learn different features and thus, achieves a competitive local learning behavior of robust low-level features.
Another problem in multi-layered SNNs is that a spiking neuron needs several spikes to emit a single spike that leads to diminishing spiking frequency \cite{tavanaei2019deep}. To tackle this problem, we use a layer-wise learning procedure where after learning the first layer is complete, its cross-depth inhibition is disabled while keeping its conductance matrix fixed. Hence, the spiking threshold for the neurons in the first layer is lowered to provide a higher spiking frequency. In this way, the neurons in the first layer receive input from the images and produce spikes that in turn facilitate the learning in the second layer and so on.


\subsection{Learning Methodology}

\begin{figure}
    \centering
    \includegraphics[width = 0.85\columnwidth]{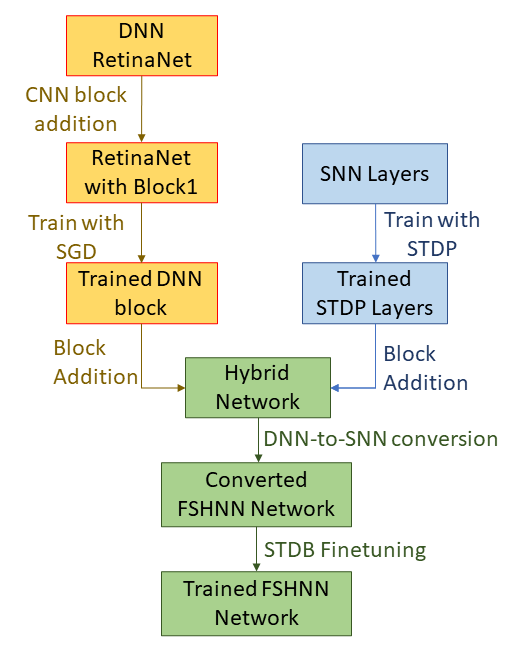}
    \caption{Flowchart Showing the Learning Methodology of the FSHNN Network}
    \label{fig:learning}
\end{figure}

The learning methodology of the FSHNN is done in the following five-step process which is summarized in Fig. \ref{fig:learning}.
\begin{itemize}
    \item We train the STDP block independently as a separate neural network as shown in Fig \ref{fig:cross} on the ImageNet traffic dataset. The ImageNet traffic dataset is a subset of the ImageNet dataset consisting of only the traffic signs, cars, and pedestrians. Since STDP is an unsupervised learning method, it need not be trained on the full ImageNet dataset. This is motivated by the observations shown by Kheradpisheh et al. \cite{kheradpisheh2018stdp}.
    \item We use these pre-trained STDP layers as a supplement to the ResNet backbone as described above.
    \item With these STDP layers frozen in place, we use the DNN-to-SNN conversion process using the process described by Miquel et al\cite{miquel2021retinanet}.
    \item For the DNN-to-SNN conversion, we use a rate-encoding of the activations. For the conversion process, we assume the firing rate of the spiking neurons over a certain time window approximates the activations of the original analog neurons \cite{rueckauer2017conversion} and there is a one-to-one correspondence between the analog and spiking neurons. The dynamics of the spiking neurons are modeled using the Leaky Integrate and Fire (LIF) model.
    \item After this conversion is done, we retrain the model with the Spike Time Dependent Backpropagation (STDB) method to finetune the weights and activations obtained. Table III. shows the implementation details of the spiking neural network STDB training method.
\end{itemize}



It is to be noted here that, in this paper we use the Focal Loss \cite{lin2017focal}, which addresses the imbalance between the trivial but over-represented background class and the object classes. Focal loss (FL) uses a modulating factor $\left(1-p_{\mathrm{t}}\right)^{\gamma}$ to the cross entropy loss, with tunable focusing parameter $\gamma \geq 0$ and is defined as:
\begin{equation}
    \mathrm{FL}\left(p_{\mathrm{t}}\right)=-\left(1-p_{\mathrm{t}}\right)^{\gamma} \log \left(p_{\mathrm{t}}\right)
\end{equation}

The parameters of the LIF and the STDP dynamics used in this paper are summarized in Table II.

\begin{table}[]
\centering
\caption{Table showing the hyperparameters used for the SNN training process}
\label{tab:stdb_implement}

\begin{tabular}{|c|c|c|c|}
\hline
\textbf{Parameter}                                                 & \textbf{Value} & \textbf{Parameter}                                                                           & \textbf{Value} \\ \hline
\textit{Batch Size}                                                & 32             & \textit{Parameter $\alpha$ for STDB}                                                         & 0.3            \\ \hline
\textit{\begin{tabular}[c]{@{}c@{}}Momentum parameter \\ for the SGD optimizer\end{tabular}} & 0.95         & \textit{Parameter $\beta$ for STDB}                                                          & 0.01           \\ \hline
\textit{Epochs}                                                    & 300            & \textit{\begin{tabular}[c]{@{}c@{}}Weight Decay parameter \\ for the optimizer\end{tabular}} & 0.0005           \\ \hline
\textit{Timesteps}                                                 & 350            & \textit{\begin{tabular}[c]{@{}c@{}}Optimizer for SNN \\ Backpropagation\end{tabular}}        & SGD            \\ \hline
\end{tabular}

\end{table}

Hence, using the weights obtained as an initialization point, we retrain these other supervised spiking layers using the Spike Time Dependent Backpropagation (STDB) process \cite{rathi2020enabling}. 
STDB is a hybrid training methodology that takes the converted SNN and uses its weights and thresholds as an initialization step for spike-based backpropagation. It then performs incremental spike-timing
dependent backpropagation (STDB) on this initialized network to obtain an SNN that converges within few epochs and requires fewer time steps for input processing. STDB is performed with a novel surrogate gradient function
defined using the neuron’s spike time.
 Thus, using the DNN-to-SNN conversion, STDP based training of SCNN layers, and STDB fine-tuning, we get a fully spiking RetinaNet object detector.

\subsection{Model Uncertainty in Spiking Object Detectors}

\subsubsection{Monte Carlo Dropout for Spiking Neural Networks}

Recent work has shown that the use of dropout during training can be used as an approximation of the Monte Carlo simulation, thus giving us an estimate of the Bayesian inference done on the network \cite{gal2016dropout}. 
Lee et al \cite{lee2020enabling} introduced the dropout technique for spiking networks, though they used it only during training and not for inference. In this paper, we use the dropout during both training and inference so that inference with dropout enabled can be interpreted as an approximate Bayesian Inference in deep Gaussian processes, similar to its DNN counterpart.

\par The dropout method in SNNs differs from that in the standard DNNs where each epoch of training has several iterations of mini-batches and during each iteration, randomly selected units are disconnected from the network while weighting by its posterior probability. On the other hand, in SNNs, each iteration has more than one forward propagation depending on the time length of the spike train. Thus, the output error is back-propagated and the network parameters are modified only at the last time step. 
Also, during training, the set of connected units within an iteration of mini-batch data remains the same in such a way that the same random subset of units constitutes the neural network during each forward propagation within a single iteration.
Similarly, during inference also, we take a random subset of the entire network in a single iteration and each iteration has multiple forward passes. Thus, we use multiple such forward samples and partition the individual detections into observations as discussed in ~\cite{miller2018dropout} to obtain the label uncertainty and the spatial bounding box uncertainty estimates.

\subsubsection{Epistemic Uncertainty Estimation for Object Detectors}

Object Detection is concerned with estimating a  bounding box alongside a label distribution for multiple objects in a scene. We extend the concept of Dropout Sampling as a means to perform tractable variational inference from image recognition to object detection. We employ the dropout sampling approximation method as described in \cite{gal2016dropout} to sample from the distribution of weights $p(\mathbf{W} \mid \mathbf{T})$ where $\mathbf{W}$ are the learned weights of the FSHNN detection network and $\mathbf{T}$ is the training data. We apply MC Dropout treating the object detector as a black box~\cite{miller2018dropout}. Uncertainty is then estimated as sample statistics from spatially correlated detector outputs. 

In the object detectors, we use the dropout layers after each of the convolution layers in the main ResNet block in the backbone network. This modified network is trained with the dropout layers as discussed above. With these trained layers, we use the dropout during inference which acts as a Monte Carlo Sampling technique. 
Every forward pass through the network corresponds to performing inference with a different network $\mathbf{\tilde{W}}$ which is approximately sampled from $p(\mathbf{W|T})$.

%% file: tex/perf_eval.tex
\section{Experimental Results}

In this section, we present experimental results to evaluate the performance of FSHNN. First, we evaluate the performance of FSHNN on the MS-COCO dataset \cite{lin2014microsoft}. The MS COCO dataset has 223K frames with instances from 81 different object categories,  and a training/testing split of 118K/5K.  Second, we evaluate the performance of FSHNN considering varying input noise levels and training the models on limited labeled data. Third, we compare FSHNN to other spiking and non-spiking object detectors trained using backpropagation considering object detection performance and energy efficiency. 
Finally, we also show that the object proposal of the FSHNN based object detector generalizes better for unseen data classes in comparison to the standard object detectors like RetinaNet. 
\par All the simulations were done on a computer with Intel Core i7-8700K CPU @ 3.70GHz CPU and 16GB NVIDIA GeForce GTX1080Ti GPU. RetinaNet trained
from scratch and FSHNN trained with hybrid conversion-and-STDB are evaluated for 100-time steps.
One epoch of training (inference) of the RetinaNet takes 79 (0.02) minutes and 8.67 (2.25) GB of GPU memory. On the other hand, one epoch of SNN training (inference) takes 125 (23.47) minutes and 15.36 (2.41) GB of GPU memory for the same hardware and batch size of 32. We evaluate the performance of the FSHNN based object detector trained for the MS-COCO dataset.

The hyperaparameters of the SNN used for the experiments are shown in Table \ref{tab:stdb_implement} and that of the LIF dynamics in Table \ref{tab:stdpparams}.

The remainder of this section is organized as follows:
\begin{enumerate}[label=(\Alph*)]
    \item \textbf{Evaluation Metrics:} Here we define the metrics used for the evaluation of the different models described
    \item \textbf{Performance Evaluation of FSHNN: }
    Here we evaluate the performance of the FSHNN models. To test the robustness of the model, we further experiment with the following adversarial conditions:
    \begin{itemize}
        \item \textit{Performance with Input Noise in Testing Images: } We evaluate the performance of the FSHNN object detector after adding different levels of Gaussian white noise to the images.
        \item \textit{Performance after Training with Less Labeled Data: } To further evaluate the performance of the FSHNN model, we re-train it with decreasing amount of labeled data and evaluated its performance.
    \end{itemize}
    \item \textbf{Comparison with Baselines: } We compare the performance of the proposed FSHNN model with some baseline object detector models. This comparison is further divided into two parts:
    \begin{itemize}
        \item \textit{Performance Comparison: } We compare the performance of the FSHNN based object detector with the other baselines.
        \item \textit{Generalizability of Object Proposals: }We perform empirical analysis to test the generalizability of the object proposal between the FSHNN and the other baselines
    \end{itemize}
    \item \textbf{Comparison with other Hybrid Models: }We compared the performance and generalizability of the FSHNN model for other DNN-based Hybrid object detection models
    \item \textbf{Ablation Studies: }We performed some ablation studies to study the significance of the addition of the STDP and the backpropagation blocks to the Spiking RetinaNet model 
    \item \textbf{Energy Efficiency Comparison: }We performed an energy efficiency of the SNN based FSHNN object detector in comparison to DNN based object detection models.
    \item \textbf{Comparison to Spiking Yolo: }We also performed an empirical comparison of the FSHNN based object detector with the Spiking YOLO model \cite{kim2020spiking}.
\end{enumerate}

\subsection{\textbf{Evaluation Metrics}}

The two main evaluation metrics used to quantify the performance of object detectors discussed above are:

\begin{itemize}

    \item  \textit{Mean Average Precision $(m A P):$} mAP is a metric for evaluating object detection performance. $\mathrm{mAP}$ measures a detector's ability to detect all objects in a closed set dataset with a correct classification and accurate localization $(\mathrm{IoU} \geq 0.5),$ while minimizing incorrect detections and their confidence score. For performance on the detection task, we use the Mean Average Precision (mAP) at 0.5 IOU. The maximum mean average precision achievable by a detector is $100 \%$. The mAP can be divided into three subclasses to depict the average precision of small object detection $AP_{S}$, medium object detection $AP_{M}$ and large object detection $AP_{L}$ depending on the size of the ground truth object. 
    
    \item \textit{Mean Average Recall $(mAR):$} Another metric for evaluating the performance of an object detector is the mean Average Recall $(mAR)$. While precision is defined by the ratio of the true positives to the sum of true positives and false positives, recall is defined as the ratio of the number of true positives to the number of ground truths. Thus, the Average Recall is defined as the recall averaged over all IoU $\in [0.5,1.0]$ and can be computed as twice the area under the recall-IoU curve. It should be noted that for the evaluation on MS-COCO, distinctions are made among different classes and its AR metric is calculated on a per-class basis, just like AP.

\item  \textit{Uncertainty Error (UE)}  \cite{miller2019evaluating}: This metric represents the ability of an uncertainty measure to accept correct detection and reject incorrect detection. The uncertainty error is the probability that detection is incorrectly accepted or rejected at a given uncertainty threshold. We use the Minimum Uncertainty Error (MUE) at 0.5 IOU to determine the ability of the detector's estimated uncertainty to discriminate true positives from false positives. The lowest MUE achievable by a detector is $0 \% .$ We define the Categorical MUE (CMUE) since we are using the Categorical entropy and finally, we average the CMUE overall categories in a testing dataset to derive the Mean (Categorical) MUE (mCMUE).
\end{itemize}

\subsection{\textbf{Performance Evaluation of FSHNN}}

In this section, we evaluate the performance of the FSHNN based object detector on the MS-COCO dataset. We do so by performing three separate experiments. First, we evaluate the performance of the model under normal circumstances, train with all the input data, and test with clean images. We repeat this experiment for multiple STDP layers in the \textbf{\texttt{Block 2}}, as shown in Fig. \ref{fig:block1}. The results are summarized in Table  \ref{tab:abl1}.  Fig. \ref{fig:fshnn} demonstrates the performance of the FSHNN object detector for a clean and noisy input image.

\begin{figure}
    \centering
    \includegraphics[width = \columnwidth]{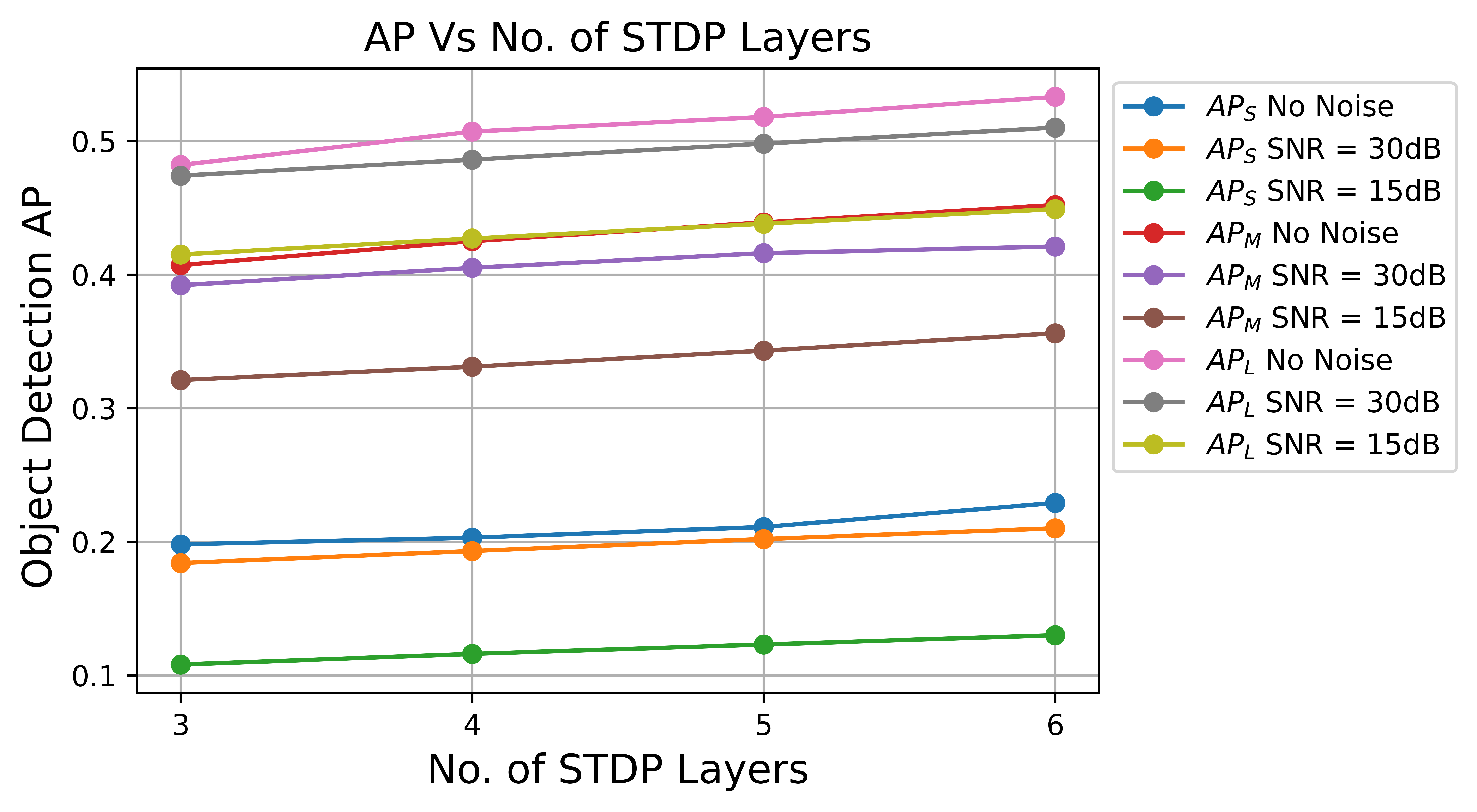}
    \caption{Figure showing the change of small, medium and large object detection AP with increasing number of STDP layers}
    \label{fig:ablation_noise}
\end{figure}

\begin{figure}
    \centering
    \includegraphics[width = \columnwidth]{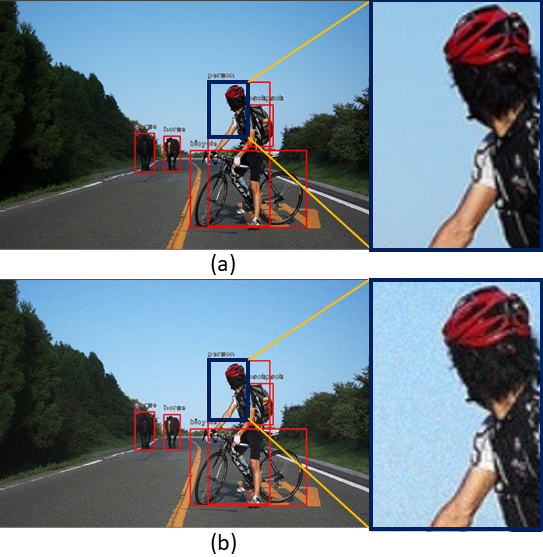}
    \caption{FSHNN object detector performance on (a) clean and (b) noisy image with added Gaussian Noise and SNR = 15dB}
    \label{fig:fshnn}
\end{figure}

We see that the insertion of more STDP layers reduces both classification and regression losses, but the classification loss reduces more. Also, from Fig. \ref{fig:ablation_noise}, we see that the inclusion of the STDP layers increases the AP for all small, medium, and large object detections.

\begin{table*}[]
\centering
\caption{Table showing Performance of the FSHNN based Object Detector with Increasing STDP Layers}
\label{tab:abl1}
\begin{tabular}{|c|c|c|c|c|c|c|c|c|c|c||c|}
\hline
\textbf{\begin{tabular}[c]{@{}c@{}}Number of \\ STDP Layers\end{tabular}} & \textbf{\begin{tabular}[c]{@{}c@{}}Classification\\  Loss\end{tabular}} & \textbf{\begin{tabular}[c]{@{}c@{}}Regression \\ Loss\end{tabular}} &  \textbf{mAP} & $\textbf{AP}_{S}$ & $\textbf{AP}_{M}$ & $\textbf{AP}_{L}$ & \textbf{mAR} & $\textbf{AR}_S$ & $\textbf{AR}_M$ & $\textbf{AR}_L$  & \textbf{mCMUE}\\ \hline 

\textbf{3}                                                               & 0.4127                                                                      & 0.6912                                                                  & 0.379  & 0.198 & 0.407 &  0.482   & 0.492    & 0.292 & 0.554 & 0.615  & 0.251 \\ \hline
\textbf{4}                                                               & 0.3841                                                                      & 0.6787                                                                  & 0.392  & 0.203 & 0.425 &  0.507  & 0.515    &  0.308 & 0.578 &  0.635 & 0.244 \\ \hline
\textbf{5}                                                               & 0.3352                                                                      & 0.6536                                                                  & 0.409    & 0.211 & 0.439 &  0.518 & 0.527   & 0.316 & 0.591 & 0.652   & 0.236 \\ \hline
\textbf{6}                                                               & 0.3076                                                                      & 0.6193                                                                  & 0.426   & 0.229 & 0.452 & 0.533  & 0.541    & 0.327 & 0.612 & 0.668 & 0.229 \\ \hline
\end{tabular}
\end{table*}


\subsubsection{Performance with Input Noise in Testing}

We evaluate the average precision and recall of the proposed FSHNN based object detector on the MS-COCO dataset under the signal-to-noise ratios (S.N.R) of 30dB and 15dB. We use a standard additive white Gaussian noise for these experiments. The results of the experiments are summarized in Table \ref{tab:abl2}. We see an increment of $12\%$ on the addition of 3 new STDP layers for the no noise case. The addition of STDP layers plays a more significant role as the network is subjected to more adversarial conditions for example when there is an increase in the input noise level or the network is trained with a lesser number of labeled training data.

\subsubsection{Performance after Training with Less Labeled Data} \label{sec:labeled}
To further demonstrate the robustness of the FSHNN object detector, we evaluated it with restrained training. 
We trained the model on a partial MS COCO dataset and reported the average precision and recall. We see that the object detectors with more STDP layers perform better especially when significant amounts of data are removed during training. Thus, we may conclude that the STDP layers trained with unlabeled data learn critical features for classification as can be seen from the decrease in classification loss with increasing STDP layers in Table \ref{tab:abl2}. To get the confidence interval of the predictions in presence of noise, we tested the trained model 5 times. The mean and variance of the mAP are shown in Table \ref{tab:comp_data}. For the case of limited training data, we performed a 3-fold cross-validation of the training subsets from the full COCO training set. The results reported are the mean and standard deviation of the mAX and the mean $AX_S , AX_M, AX_L$, mCMUE (where AX denotes AP or AR). We observe that the confidence interval is proportional to the mCMUE calculated. Also, it must be noted here that the change in the training data is for the backpropagated layers and not the STDP layers which are still trained on the full ImageNet traffic dataset as described before. However, the ImageNet traffic dataset used for training the STDP layers are unlabeled data and are thus much readily accessible compared to labeled data.

\begin{table}[]
\centering
\caption{Table showing Robustness of FSHNN based Object Detector for Noisy Testing Data and Limited Labeled Training Dataset }
\label{tab:abl2}
\begin{tabular}{|c|c|c|c|c|c|}
\hline
\textbf{\begin{tabular}[c]{@{}c@{}}\# STDP\\ Layers\end{tabular}} & \textbf{\begin{tabular}[c]{@{}c@{}}Class.\\  Loss\end{tabular}} & \textbf{\begin{tabular}[c]{@{}c@{}}Reg. \\ Loss\end{tabular}} & \textbf{\begin{tabular}[c]{@{}c@{}}mAP \\ (mean$\pm$var) \end{tabular}} & \textbf{\begin{tabular}[c]{@{}c@{}}mAR \\ (mean$\pm$var) \end{tabular}} & \textbf{\begin{tabular}[c]{@{}c@{}}mean \\ mCMUE \end{tabular}}  \\ \hline
\multicolumn{6}{|c|}{\textbf{SNR = 30dB}}                                                                                                                                                                                                                              \\ \hline
\textbf{3}                                                               & 0.4318                                                                      & 0.7165                                                                  & \begin{tabular}[c]{@{}c@{}}0.366 \\ $\pm 0.00438$\end{tabular}       & \begin{tabular}[c]{@{}c@{}}0.480 \\ $\pm 0.00473$\end{tabular}       & 0.317 \\ \hline
\textbf{4}                                                               & 0.3987                                                                      & 0.7093                                                                  & \begin{tabular}[c]{@{}c@{}}0.379 \\ $\pm 0.00419$\end{tabular}        & \begin{tabular}[c]{@{}c@{}}0.494 \\ $\pm 0.00466$\end{tabular}        & 0.305 \\ \hline
\textbf{5}                                                               & 0.3516                                                                      & 0.6845                                                                  & \begin{tabular}[c]{@{}c@{}}0.392 \\ $\pm 0.00411$\end{tabular}        & \begin{tabular}[c]{@{}c@{}}0.517 \\ $\pm 0.00462$\end{tabular}        & 0.291 \\ \hline
\textbf{6}                                                               & 0.3204                                                                      & 0.6522                                                                  & \begin{tabular}[c]{@{}c@{}}0.412 \\ $\pm 0.00403$\end{tabular}        & \begin{tabular}[c]{@{}c@{}}0.531 \\ $\pm 0.00452$\end{tabular}        & 0.284 \\ \hline
\multicolumn{6}{|c|}{\textbf{SNR = 15dB}}                                                                                                                                                                                                                              \\ \hline
\textbf{3}                                                               & 0.6723                                                                      & 0.7345                                                                  & \begin{tabular}[c]{@{}c@{}}0.302 \\ $\pm 0.00501$\end{tabular}       & \begin{tabular}[c]{@{}c@{}}0.441 \\ $\pm 0.00522$\end{tabular}        & 0.455 \\ \hline
\textbf{4}                                                               & 0.6552                                                                      & 0.7269                                                                  & \begin{tabular}[c]{@{}c@{}}0.316 \\ $\pm 0.00490$\end{tabular}        & \begin{tabular}[c]{@{}c@{}}0.453 \\ $\pm 0.00513$\end{tabular}        & 0.448 \\ \hline
\textbf{5}                                                               & 0.6218                                                                      & 0.7114                                                                  & \begin{tabular}[c]{@{}c@{}}0.321 \\ $\pm 0.00488$\end{tabular}        & \begin{tabular}[c]{@{}c@{}}0.467 \\ $\pm 0.00511$\end{tabular}        & 0.439 \\ \hline
\textbf{6}                                                               & 0.6014                                                                      & 0.7008                                                                  & \begin{tabular}[c]{@{}c@{}}0.330 \\ $\pm 0.00485$\end{tabular}        & \begin{tabular}[c]{@{}c@{}}0.477 \\ $\pm 0.00503$\end{tabular}        & 0.431 \\ \hline
\hline
\multicolumn{6}{|c|}{\textbf{80\% of Labeled Training Data}}                                                                                                                                                                                                                               \\ \hline
\textbf{3}                                                               & 0.4157                                                                      & 0.6998                                                                  & \begin{tabular}[c]{@{}c@{}}0.358 \\ $\pm 0.00453$\end{tabular}        & \begin{tabular}[c]{@{}c@{}}0.475 \\ $\pm 0.00482$\end{tabular}       & 0.301 \\ \hline
\textbf{4}                                                               & 0.3806                                                                      & 0.6742                                                                  & \begin{tabular}[c]{@{}c@{}}0.362 \\ $\pm 0.00450$\end{tabular}        & \begin{tabular}[c]{@{}c@{}}0.480 \\ $\pm 0.00479$\end{tabular}        & 0.289 \\ \hline
\textbf{5}                                                               & 0.3694                                                                      & 0.6516                                                                  & \begin{tabular}[c]{@{}c@{}}0.390 \\ $\pm 0.00445$\end{tabular}        & \begin{tabular}[c]{@{}c@{}}0.513 \\ $\pm 0.00473$\end{tabular}        & 0.214 \\ \hline
\textbf{6}                                                               & 0.3356                                                                      & 0.6288                                                                  & \begin{tabular}[c]{@{}c@{}}0.412 \\ $\pm 0.00439$\end{tabular}        & \begin{tabular}[c]{@{}c@{}}0.547 \\ $\pm 0.00467$\end{tabular}        & 0.199 \\ \hline
\multicolumn{6}{|c|}{\textbf{60\% of Labeled Training Data}}                                                                                                                                                                                                                               \\ \hline
\textbf{3}                                                               & 0.4975                                                                      & 0.7884                                                                  & \begin{tabular}[c]{@{}c@{}}0.329 \\ $\pm 0.00482$\end{tabular}        & \begin{tabular}[c]{@{}c@{}}0.435 \\ $\pm 0.00501$\end{tabular}        & 0.390 \\ \hline
\textbf{4}                                                               & 0.4698                                                                      & 0.7545                                                                  & \begin{tabular}[c]{@{}c@{}}0.341 \\ $\pm 0.00481$\end{tabular}        & 0\begin{tabular}[c]{@{}c@{}}0.476 \\ $\pm 0.00497$\end{tabular}        & 0.358 \\ \hline
\textbf{5}                                                               & 0.4513                                                                      & 0.7312                                                                  & \begin{tabular}[c]{@{}c@{}}0.364 \\ $\pm 0.00484$\end{tabular}        & \begin{tabular}[c]{@{}c@{}}0.490 \\ $\pm 0.00499$\end{tabular}        & 0.317 \\ \hline
\textbf{6}                                                               & 0.4027                                                                      & 0.7084                                                                  & \begin{tabular}[c]{@{}c@{}}0.391 \\ $\pm 0.00482$\end{tabular}        & \begin{tabular}[c]{@{}c@{}}0.512 \\ $\pm 0.00496$\end{tabular}        & 0.296 \\ \hline
\multicolumn{6}{|c|}{\textbf{40\% of Labeled Training Data}}                                                                                                                                                                                                                               \\ \hline
\textbf{3}                                                               & 0.6057                                                                      & 0.9432                                                                  & \begin{tabular}[c]{@{}c@{}}0.285 \\ $\pm 0.00520$\end{tabular}        & \begin{tabular}[c]{@{}c@{}}0.385 \\ $\pm 0.00553$\end{tabular}        & 0.513 \\ \hline
\textbf{4}                                                               & 0.5964                                                                      & 0.9303                                                                  & \begin{tabular}[c]{@{}c@{}}0.294 \\ $\pm 0.00516$\end{tabular}        & \begin{tabular}[c]{@{}c@{}}0.400 \\ $\pm 0.00550$\end{tabular}        & 0.501 \\ \hline
\textbf{5}                                                               & 0.5837                                                                      & 0.9268                                                                  & \begin{tabular}[c]{@{}c@{}}0.299 \\ $\pm 0.00511$\end{tabular}        & \begin{tabular}[c]{@{}c@{}}0.415 \\ $\pm 0.00546$\end{tabular}       & 0.486 \\ \hline
\textbf{6}                                                               & 0.5759                                                                      & 0.9115                                                                  & \begin{tabular}[c]{@{}c@{}}0.319 \\ $\pm 0.00504$\end{tabular}        & \begin{tabular}[c]{@{}c@{}}0.423 \\ $\pm 0.00541$\end{tabular}        & 0.478 \\ \hline
\end{tabular}
\end{table}

\subsection{\textbf{Comparison with Baselines}}

 We compare the performance and energy efficiency of FSHNN with different baselines. The two baseline models for this comparison are described as follows (Table~\ref{tab:layers}):
\begin{itemize}
    \item \textit{RetinaNet: } RetinaNet is a state-of-the-art DNN based object detector that works using the Focal Pyramid Network architecture. In this paper, we are using ResNet 101 as the backbone network for the RetinaNet object detector. 
        \item \textit{Back Propagated Fully Spiking Object Detector (BPSOD): } For the backpropagated fully spiking RetinaNet we replace the STDP layers (\texttt{Block 1} in Fig. 1)
        in the FSHNN based object detector with corresponding DNN layers. For this experiment, we have replaced 3 STDP layers with 3 Convolution Layers. Hence, we use a similar DNN-to-SNN conversion methodology to obtain a Fully Spiking Object Detector with backpropagated spiking layers with the same architecture as FSHNN based object detector. With the weights obtained in this conversion process as the initialization point, we retrain the model using STBP to get the final trained model. 
\end{itemize}
\begin{table}[h]
\centering

\caption{Table Describing the Learning Blocks used in the Architectures}
\label{tab:layers}
\resizebox{0.45\textwidth}{!}{%
\begin{tabular}{|c|c|c|c|c|c|}
\hline
 &  & \textbf{\texttt{Block 1}} & \textbf{\texttt{Block 2}} & \textbf{\begin{tabular}[c]{@{}c@{}}ResNet\\ Block\end{tabular}} & \textbf{\begin{tabular}[c]{@{}c@{}}FPN\\  Block\end{tabular}} \\ \hline
\multirow{2}{*}{\textbf{RetinaNet}} & Layer & None & None & DNN & DNN \\ \cline{2-6} 
 & Learning & N/A & N/A & BP & BP \\ \hline
\multirow{2}{*}{\textbf{\begin{tabular}[c]{@{}c@{}}FSHNN \\ Obj Det\end{tabular}}} & Layer & SNN & SNN & SNN & SNN \\ \cline{2-6} 
 & Learning & BP & STDP & BP & BP \\ \hline

\multirow{2}{*}{\textbf{BPSOD}} & Layer & SNN & SNN & SNN & SNN \\ \cline{2-6} 
 & Learning & BP & BP & BP & BP \\ \hline
\end{tabular}%
}
\end{table}

The parameters used for the RetinaNet and the BPSOD are summarized in Table \ref{tab:base_params}.

\begin{table}[]
\centering
\caption{Table showing the hyperparameters used for the RetinaNet and BPSOD object detectors}
\label{tab:base_params}

\begin{tabular}{|c|c||c|c|}
\hline
\textbf{\begin{tabular}[c]{@{}c@{}}RetinaNet \\ Parameters \end{tabular}}                                                 & \textbf{Value} & \textbf{\begin{tabular}[c]{@{}c@{}}BPSOD \\ Parameters \end{tabular}}                                                                          & \textbf{Value} \\ \hline
\textit{Batch Size}                                                & 32             & \textit{Batch Size}                                                         & 32         \\ \hline

\textit{Epochs} & 100 & \textit{Epochs} & 100 \\ \hline

\textit{\begin{tabular}[c]{@{}c@{}} Backbone \\ Size \end{tabular}} & \begin{tabular}[c]{@{}c@{}}ResNet- \\ 101\end{tabular} & \textit{\begin{tabular}[c]{@{}c@{}} Max to Average \\ Pooling \end{tabular}}                                                          & \texttt{True}           \\ \hline
\textit{Optimizer}                                                    & SGD            & \textit{\begin{tabular}[c]{@{}c@{}}Spike \\ Coding \end{tabular}} & \begin{tabular}[c]{@{}c@{}}Temporal Mean \\ Rate\end{tabular}            \\ \hline
\textit{\begin{tabular}[c]{@{}c@{}} Learning\\ Rate \end{tabular}}                                                 & $0.001$            & \textit{\begin{tabular}[c]{@{}c@{}}MaxPool\\ Type\end{tabular}}        & \begin{tabular}[c]{@{}c@{}}Accumulated absolute\\ spike rate \cite{hu2016max} \end{tabular}            \\ \hline
\end{tabular}
\end{table}

\begin{table*}[]
\centering
\caption{Table showing the performance of the different object detectors}
\label{tab:perf_comp}
\resizebox{0.65\textwidth}{!}{%
\begin{tabular}{|c|c|c|c|c|c|c|c|c||c|}
\hline

& \textbf{mAP } & $\textbf{AP}_{S}$ & $\textbf{AP}_{M}$ & $\textbf{AP}_{L}$ & \textbf{mAR} & $\textbf{AR}_S$ & $\textbf{AR}_M$ & $\textbf{AR}_L$  & \textbf{mCMUE}\\ \hline
\multicolumn{10}{|c|}{\textbf{No Noise; 100\% Training Data}}                                                                                                                                                           \\ \hline
\textbf{RetinaNet}                                                          & 0.388       & 0.207        & 0.420        & 0.498        & 0.501       & 0.301        & 0.566        & 0.624    &  0.237    \\ \hline

\textbf{BPSOD} & 0.375       & 0.189        & 0.401        & 0.477        & 0.486       & 0.283        & 0.549        & 0.598      &  0.254   \\ \hline
\textbf{\begin{tabular}[c]{@{}c@{}}FSHNN Obj Det \\  (6 STDP layers)\end{tabular}}                                                      &    0.426    &     0.223    &   0.452      &   0.526      &   0.541     &   0.312      &    0.611     &   0.682   &  0.229  \\ \hline
\end{tabular}%
}
\end{table*}

\begin{table*}[]
\centering

\caption{Table showing the performance of the different object detectors with input noise during testing}
\label{tab:comp_noise}
\resizebox{0.65\textwidth}{!}{%
\begin{tabular}{|c|c|c|c|c|c|c|c|c||c|}
\hline

& \textbf{\begin{tabular}[c]{@{}c@{}}mAP \\ (mean$\pm$var) \end{tabular}} & $\textbf{AP}_{S}$ & $\textbf{AP}_{M}$ & $\textbf{AP}_{L}$ & \textbf{\begin{tabular}[c]{@{}c@{}}mAR \\ (mean$\pm$var) \end{tabular}} & $\textbf{AR}_S$ & $\textbf{AR}_M$ & $\textbf{AR}_L$  & \textbf{\begin{tabular}[c]{@{}c@{}}mean \\ mCMUE \end{tabular}}\\ \hline

\multicolumn{10}{|c|}{\textbf{SNR = 30dB}}                                                                                                                                                         \\ \hline
\textbf{RetinaNet}                                                          & \begin{tabular}[c]{@{}c@{}}0.372 \\ $\pm 0.00419$\end{tabular}         & 0.183        & 0.409        & 0.481        & \begin{tabular}[c]{@{}c@{}}0.489 \\ $\pm 0.00439$\end{tabular}         & 0.279        & 0.554        & 0.613   & 0.313  \\ \hline

\textbf{BPSOD} & \begin{tabular}[c]{@{}c@{}}0.363 \\ $\pm 0.00425$\end{tabular}         & 0.186        & 0.388        & 0.472        & \begin{tabular}[c]{@{}c@{}}0.475 \\ $\pm 0.00443$\end{tabular}         & 0.278        & 0.543        & 0.589       & 0.320 \\ \hline
\textbf{\begin{tabular}[c]{@{}c@{}}FSHNN Obj Det \\  (6 STDP layers)\end{tabular}}                                                      &  \begin{tabular}[c]{@{}c@{}}0.409 \\ $\pm 0.00409$\end{tabular}        &   0.211      &   0.446      &       0.482  &     \begin{tabular}[c]{@{}c@{}}0.528 \\ $\pm 0.00430$\end{tabular}     &  0.308       &  0.599        &    0.668    &  0.284  \\ \hline
\multicolumn{10}{|c|}{\textbf{SNR = 15dB}}                                                                                                                                                         \\ \hline
\textbf{RetinaNet}                                                          & \begin{tabular}[c]{@{}c@{}}0.274 \\ $\pm 0.00611$\end{tabular}         & 0.104        & 0.237        & 0.349        & \begin{tabular}[c]{@{}c@{}}0.412 \\ $\pm 0.00630$\end{tabular}         & 0.204        & 0.401        & 0.426       & 0.605  \\ \hline

\textbf{BPSOD} & \begin{tabular}[c]{@{}c@{}}0.288 \\ $\pm 0.00605$\end{tabular}         & 0.159        & 0.347        & 0.406        & \begin{tabular}[c]{@{}c@{}}0.419 \\ $\pm 0.00623$\end{tabular}         & 0.248        & 0.473        & 0.539       & 0.483  \\ \hline
\textbf{\begin{tabular}[c]{@{}c@{}}FSHNN Obj Det \\  (6 STDP layers)\end{tabular}}                                                    &   \begin{tabular}[c]{@{}c@{}}0.328 \\ $\pm 0.00501$\end{tabular}       &   0.184      &    0.392     &   0.451      &    \begin{tabular}[c]{@{}c@{}}0.474 \\ $\pm 0.00527$\end{tabular}       &    0.271    &    0.528     &  0.587    &  0.431  \\ \hline

\end{tabular}%
}

\end{table*}

\begin{table*}[]
\centering
\caption{Table showing the performance of the different object detectors with different amount of labeled data}

\label{tab:comp_data}
\resizebox{0.65\textwidth}{!}{%
\begin{tabular}{|c|c|c|c|c|c|c|c|c||c|}
\hline
& \textbf{\begin{tabular}[c]{@{}c@{}}mAP \\ (mean$\pm$var) \end{tabular}} & $\textbf{AP}_{S}$ & $\textbf{AP}_{M}$ & $\textbf{AP}_{L}$ & \textbf{\begin{tabular}[c]{@{}c@{}}mAR \\ (mean$\pm$var) \end{tabular}} & $\textbf{AR}_S$ & $\textbf{AR}_M$ & $\textbf{AR}_L$  & \textbf{\begin{tabular}[c]{@{}c@{}}mean \\ mCMUE \end{tabular}}\\ \hline
\multicolumn{10}{|c|}{\textbf{80\% of Labeled training set}}                                                                                                                                                  \\ \hline
\textbf{RetinaNet}                                                          & \begin{tabular}[c]{@{}c@{}}0.369 \\ $\pm 0.00423$\end{tabular}        & 0.189        & 0.398        & 0.478        & \begin{tabular}[c]{@{}c@{}}0.480 \\ $\pm 0.00435$\end{tabular}        & 0.287        & 0.543        & 0.608     &  0.308  \\ \hline

\textbf{BPSOD} & \begin{tabular}[c]{@{}c@{}}0.349 \\ $\pm 0.00416$\end{tabular}        & 0.178        & 0.379        & 0.460        & \begin{tabular}[c]{@{}c@{}}0.465 \\ $\pm 0.00456$\end{tabular}       & 0.264        & 0.525        & 0.574      &   0.305 \\ \hline
\textbf{\begin{tabular}[c]{@{}c@{}}FSHNN Obj Det \\  (6 STDP layers)\end{tabular}}                                                      & \begin{tabular}[c]{@{}c@{}}0.408 \\ $\pm 0.00401$\end{tabular}       & 0.221        & 0.452        & 0.521        & \begin{tabular}[c]{@{}c@{}}0.542 \\ $\pm 0.00421$\end{tabular}       & 0.311        & 0.594        & 0.662      & 0.199   \\ \hline
\multicolumn{10}{|c|}{\textbf{60\% of Labeled training set}}                                                                                                                                                  \\ \hline
\textbf{RetinaNet}                                                          & \begin{tabular}[c]{@{}c@{}}0.322 \\ $\pm 0.00488$\end{tabular}       & 0.137        & 0.342        & 0.431        & \begin{tabular}[c]{@{}c@{}}0.431 \\ $\pm 0.00515$\end{tabular}       & 0.231        & 0.492        & 0.557      &  0.393  \\ \hline

\textbf{BPSOD} & \begin{tabular}[c]{@{}c@{}}0.313 \\ $\pm 0.00427$\end{tabular}       & 0.129        & 0.328        & 0.424        & \begin{tabular}[c]{@{}c@{}}0.420 \\ $\pm 0.00469$\end{tabular}       & 0.221        & 0.476        & 0.541      &  0.396  \\ \hline
\textbf{\begin{tabular}[c]{@{}c@{}}FSHNN Obj Det \\  (6 STDP layers)\end{tabular}}                                                     &   \begin{tabular}[c]{@{}c@{}}0.392 \\ $\pm 0.00402$\end{tabular}     &    0.188     &   0.409      &   0.496      &    \begin{tabular}[c]{@{}c@{}}0.518 \\ $\pm 0.00436$\end{tabular}    &   0.271      &    0.542     &   0.608   & 0.296   \\ \hline
\multicolumn{10}{|c|}{\textbf{40\% of Labeled training set}}                                                                                                                                                  \\ \hline
\textbf{RetinaNet}                                                          & \begin{tabular}[c]{@{}c@{}}0.236 \\ $\pm 0.00653$\end{tabular}       & 0.056        & 0.266        & 0.351        & \begin{tabular}[c]{@{}c@{}}0.345 \\ $\pm 0.00679$\end{tabular}       & 0.158        & 0.413        & 0.472      & 0.799  \\ \hline

\textbf{BPSOD} & \begin{tabular}[c]{@{}c@{}}0.251 \\ $\pm 0.00617$\end{tabular}       & 0.076        & 0.262        & 0.368        & \begin{tabular}[c]{@{}c@{}}0.362 \\ $\pm 0.00632$\end{tabular}       & 0.163        & 0.412        & 0.487     &  0.613  \\ \hline
\textbf{\begin{tabular}[c]{@{}c@{}}FSHNN Obj Det \\  (6 STDP layers)\end{tabular}}                                                     &   \begin{tabular}[c]{@{}c@{}}0.319 \\ $\pm 0.00506$\end{tabular}     &   0.126      &   0.337      &  0.439      &       \begin{tabular}[c]{@{}c@{}}0.421 \\ $\pm 0.00529$\end{tabular}  &    0.221    &   0.492      &  0.553  & 0.478   \\ \hline
\end{tabular}%
}
\end{table*}

\subsubsection{Performance Comparison}

We perform a two-fold evaluation of the FSHNN model with the baselines described above -  first we evaluate the performance of the object detectors without any perturbation and train with the full training dataset. The results are represented in Table \ref{tab:perf_comp}. We see that the FSHNN with 6 STDP layers outperforms the RetinaNet with respect to both mAP/mAR and uncertainty error.

We test the robustness of the model by testing the model with different adversarial conditions - first, we use two different noise levels (SNR = 30dB, 15dB) and also evaluate the performance after training the model with increasingly limited labeled data (80\%, 60\%, and 40\% data respectively).

 We observe that in the no noise scenario, the FSHNN with 6 STDP layers outperforms the RetinaNet, and the difference in their performances increases as the input noise level is increased. Further, when the models are trained with a lesser amount of labeled data, even the FSHNN with 3 STDP layers perform better than the baseline ones. 
 
In addition to this, we retrained and re-evaluated the models with varying conditions to get the confidence interval of the predictions as described in Sec. \ref{sec:labeled}. We report the results for the mean and variance of mAX and the mean $AX_S , AX_M, AX_L$, mCMUE (where AX denotes AP or AR). The baseline performance of the model and the performance in presence of testing noise are shown in Tables \ref{tab:perf_comp} and \ref{tab:comp_noise} respectively. The model's performance with limited training data is tabulated in Table \ref{tab:comp_data}.

From Table \ref{tab:perf_comp}, we see that the FSHNN based object detector outperforms BPSOD.  Thus, the use of STDP trained layers in the SNN shows a significant improvement in the mAP performance of the object detector compared to the backpropagated layers. This is because the STDP learning process extracts features that normal backpropagation-based feature extractors are unable to detect \cite{she2020safe}, \cite{kheradpisheh2018stdp}, especially its ability to detect small objects as seen in Tables \ref{tab:perf_comp}, \ref{tab:comp_noise}, \ref{tab:comp_data}. Thus, the addition of STDP layers can be thought of as including a specialized addition of features that the BPSOD cannot capture and thus enhancing the performance. 
To empirically prove this, we evaluated the training loss of the BPSOD and the FSHNN object detectors. After 100 epochs, the FSHNN with 3-layer STDP had a training loss of $0.137$, FSHNN with 6 STDP layers have a training loss of $0.129$ and the BPSOD has a training loss of $0.162$. Since the BPSOD has a higher training loss and also a higher testing loss, it shows that BPSOD is unable to extract features that the STDP based FSHNN can. 
The variation of training parameters does not lead to a huge variance in the observed mAP and the mAR of the BPSOD as can be seen in Table \ref{tab:comp_data}. Thus, we can conclude that the BPSOD does not face the problem of converging to a local minima. Hence, we conclude that the FSHNN based model outperforms BPSOD  because of better extraction of local features.

\subsubsection{Generalizability of Object Proposal} 

As defined by Wang et. al, the generalizability of an object detector model is its ability to localize (not classify) unannotated objects in the training dataset  \cite{wang2020leads}. For this experiment, we randomly split the MS-COCO dataset into two parts - the source dataset consisting of 70 seen classes and the target dataset of 10 unseen classes.  We use the target dataset to evaluate the generalization of the proposal model trained with the source dataset. The train split is utilized for training and the 5000 images from the validation set during evaluation. 

To evaluate the quality of the proposals, we use the standard average recall (AR@k) \cite{hosang2015makes}. One of the primary motivations for building a generalized proposal model is to use the resulting proposals to train detection models for unseen classes with limited or no bounding box annotation.

We compare the generalization ability of the FSHNN based object detector and RetinaNet in Table. \ref{tab:gen1}. The models are trained on the COCO-source-train dataset as described above. We report AR@100 on seen classes in the COCO-source-test dataset and unseen classes in the COCO-target-test. We build on the hypothesis that the difference in performance between seen and unseen classes reflects the generalization gap. We also show an upper-bound performance on COCO-target-test obtained by models trained on the full training dataset containing both COCO-source-train and COCO-target-train. On seen classes, RetinaNet achieves a worse performance compared to FSHNN (a drop of 7.01\%). However, the drop is larger for unseen target classes (a drop of 11.9\%), indicating a larger generalization gap for RetinaNet. One reason for this is that RetinaNet is more sensitive to missing bounding boxes corresponding to unlabeled unseen classes in the source dataset. Also, since RetinaNet uses focal-loss, the unseen and unannotated object classes proposals in the training data are treated as hard-negatives.  Thus, the model heavily penalizes proposals corresponding to unannotated bounding boxes, leading to an overall decrease in AR. Since some seen classes share visual similarities with unseen classes, this sensitivity to missing annotations also affects AR for seen classes.  However,  this effect is more magnified for unseen target classes.  
On the other hand, in FSHNN, only a small number of proposals that do not intersect with annotated bounding boxes are sampled at random as negatives. Hence, the probability that a proposal corresponding to an unseen object class is negative is lower, leading to better generalization. 
Thus, we observe that the detection head of  FSHNN  provides better overall performance without sacrificing generalization.


\begin{table}[]
\centering
\caption{AR@100  corresponding  to  different baseline models  trained on  COCO-source-train  and  evaluated  on  different  test  splits. Upper-bound  corresponds  to  model  trained  on  full  COCO dataset and evaluated on COCO-target-test.}

\begin{tabular}{|c|c|c|}
\hline
                                                                        & \textbf{COCO-source-test} & \textbf{COCO-target-test} \\ \hline
\textbf{RetinaNet}                                                               & 55.7                      & 41.5                      \\ \hline
\textbf{BPSOD}            & 51.8                      & 39.9                      \\ \hline
\textbf{\begin{tabular}[c]{@{}c@{}}FSHNN Obj Det\\ (3 STDP Layers)\end{tabular}} & 55.7                      & 47.9                      \\ \hline
\textbf{\begin{tabular}[c]{@{}c@{}}FSHNN Obj Det\\ (6 STDP Layers)\end{tabular}} & 56.2                      & 48.4                      \\ \hline
\end{tabular}

\label{tab:gen1}
\end{table}

\subsection{Comparison with other Hybrid Models} \label{sec:hybrid}

In this section, we compare the performance of the FSHNN based object detectors with other hybrid models. To do so, we replace the STDP based Spiking CNN layers (\texttt{Block 2}) shown in Fig. \ref{fig:block1} with other unsupervised feature extractors and evaluated the new object detectors obtained on the MS-COCO dataset similar to the FSHNN model. It is to be noted here that the rest of the models are the same as FSHNN i.e., they are trained with a DNN-to-SNN conversion followed by a STDB finetuning, as shown in Fig. \ref{fig:learning}. Also, it is to be noted here that the unsupervised models mentioned here are based on DNNs and not converted to SNNs using conversion techniques. The features extracted using these methods are then converted into spikes using a Difference of Gaussian (DoG) filter \cite{kheradpisheh2018stdp} which generates a spike train detecting the positive or negative contrasts in the feature space. The output of the DoG filter is fed into the spiking ResNet block after concatenating with the spiking output of \texttt{Block 1}. The unsupervised models used for the comparison are listed as follows:

\begin{itemize}
    \item \textbf{Unsupervised Convolutional Siamese Network (UCSN) :} We use an Unsupervised Convolutional Siamese Network for feature extraction \cite{trosten2019unsupervised}. It is a new deep learning-based feature extractor consisting of an end-to-end trained convolutional neural network for adaptive neighborhood embedding in an unsupervised manner.
    In this experiment, we use the UCSN-HOG which uses the distance between the Histogram of Oriented Gradients (HOG) features which is a type of global image descriptor. 

    \item \textbf{Convolutional Autoencoder (CAE):} They are a type of Convolutional Neural Networks (CNNs) that are trained only to learn filters able to extract features that can be used to reconstruct the input. CNNs are usually referred to as supervised learning algorithms. The latter, instead, are trained only to learn filters able to extract features that can be used to reconstruct the input. Each CAE is trained using conventional online gradient descent without additional regularization terms. For the simulation, we initialized a convolutional neural network (CNN) with trained convolutional auto-encoder weights as described by Masci et al. \cite{masci2011stacked}. 
\end{itemize}
The detailed block architecture are summarized in Table \ref{tab:layers_hybrid} and the results are shown in Table \ref{tab:hybrid_perf}. We see that the FSHNN based object detector outperforms the other DNN based hybrid learning models, especially in small object detection.

\textbf{Generalizability of Hybrid Networks: } Similarly, we also check the generalizability of the other hybrid models discussed above. We evaluate these hybrid models on the COCO-source-test and the COCO-target-test datasets and reported their AR@100 scores in Table \ref{tab:gen2}. We see that the generalizability of object proposals of the FSHNN based object detector is better than the DNN based unsupervised hybrid models. This shows that STDP based feature extractors generalize better than alternative unsupervised learning models for DNN.

\begin{table}[]
\centering

\caption{Table Describing the Learning Blocks used in the \textbf{Other Hybrid Learning Architectures} for Object Detection}
\label{tab:layers_hybrid}
\resizebox{0.5\textwidth}{!}{%
\begin{tabular}{|c|c|c|c|c|c|}
\hline
 &  & \textbf{\texttt{Block 1}} & \textbf{\texttt{Block 2}} & \textbf{\begin{tabular}[c]{@{}c@{}}ResNet\\ Block\end{tabular}} & \textbf{\begin{tabular}[c]{@{}c@{}}FPN\\  Block\end{tabular}} \\ \hline
   \multirow{2}{*}{\textbf{\begin{tabular}[c]{@{}c@{}} Hybrid UCSN \\ Obj Det\end{tabular}}} & Layer & SNN & DNN & SNN & SNN \\ \cline{2-6} 
 & Learning & BP & UCSN & BP & BP \\ \hline
   \multirow{2}{*}{\textbf{\begin{tabular}[c]{@{}c@{}} Hybrid CAE \\ Obj Det\end{tabular}}} & Layer & SNN & DNN & SNN & SNN \\ \cline{2-6} 
 & Learning & BP & CAE & BP & BP \\ \hline
\end{tabular}%
}
\end{table}

\begin{table*}[]
\centering

\caption{Table showing Performance Comparison of Object Detectors based on \textbf{other Hybrid Models}}
\label{tab:hybrid_perf}
\begin{tabular}{|c|c|c|c|c|c|c|c|c|}
\hline
\textbf{\begin{tabular}[c]{@{}c@{}}Number of \\ STDP Layers\end{tabular}} &  \textbf{mAP} & $\textbf{AP}_{S}$ & $\textbf{AP}_{M}$ & $\textbf{AP}_{L}$ & \textbf{mAR} & $\textbf{AR}_S$ & $\textbf{AR}_M$ & $\textbf{AR}_L$  \\ \hline 
\textbf{\begin{tabular}[c]{@{}c@{}}FSHNN Obj Det \\ (6 STDP Layers) \end{tabular}}                                                                                                                    & 0.426   & 0.229 & 0.452 & 0.533  & 0.541    & 0.327 & 0.612 & 0.668  \\ \hline
\textbf{Hybrid UCSN Obj Det}                                                                                                                   & 0.383   & 0.191 & 0.411 & 0.487  & 0.497    & 0.290 & 0.559 & 0.621  \\ \hline
\textbf{ Hybrid UCSN CAE Det}                                                                                                                   & 0.381   & 0.189 & 0.415 & 0.486  & 0.500    & 0.289 & 0.561 & 0.627  \\ \hline
\end{tabular}
\end{table*}

\begin{table}[]
\centering
\caption{AR@100  corresponding  to  different hybrid models  trained on  COCO-source-train  and  evaluated  on  different  test  splits.}

\begin{tabular}{|c|c|c|}
\hline
                                                                                 & \textbf{COCO-source-test} & \textbf{COCO-target-test} \\ \hline
\textbf{\begin{tabular}[c]{@{}c@{}}Hybrid UCSN\\ Obj Det\end{tabular}}           & 51.8                      & 36.7                      \\ \hline
\textbf{\begin{tabular}[c]{@{}c@{}}Hybrid CAE\\ Obj Det\end{tabular}}            & 51.4                      & 36.4                      \\ \hline
\textbf{\begin{tabular}[c]{@{}c@{}}FSHNN Obj Det\\ (6 STDP Layers)\end{tabular}} & 56.2                      & 48.4                      \\ \hline
\end{tabular}
\label{tab:gen2}
\end{table}

\subsection{Ablation Studies } 
We consider the ablation studies of the FSHNN object detector discussed before. 

The three types of ablation network studied in this section are described as follows:
\begin{itemize}
    \item \textbf{STDP-STDP Spiking Object Detector: }This object detector is made using a 4-layer STDP trained SNN block in place of the Backpropagated SCNN module (\texttt{Block 1} in Fig. \ref{fig:block1}) in addition to the STDP SCNN module (\texttt{Block 2} in Fig. \ref{fig:block1})
    \item \textbf{Null-STDP Spiking Object Detector: }This ablation network is obtained by removing the BP SCNN module (\texttt{Block 1}) from the FSHNN model and adding only the STDP SCNN module (\texttt{Block 2}) to the Spiking RetinaNet model.
    \item \textbf{BP-Null Spiking Object Detector: } For this ablation study, we remove the STDP SCNN module (\texttt{Block 2}) from the FSHNN model and only keep the BP SCNN module (\texttt{Block 1}). 
\end{itemize}
We evaluate the performance of these ablation networks on the MS-COCO dataset. The detailed block architecture is summarized in Table \ref{tab:layers_abl} and the results are shown in Table \ref{tab:perf_abl}.  We see that the FSHNN object detector outperforms the other ablation networks. This ablation study also proves that the sole usage of either STDP based layers or back-propagated spiking layers cannot extract the features which a hybrid method can. 

\textbf{Generalizability of Ablation Networks: } Similarly, we also check the generalizability of the ablation networks discussed above. We also evaluate these ablation models on the COCO-source-test and the COCO-target-test datasets and evaluated their AR@100 scores. The results are also shown in Table \ref{tab:gen3}. We see that the generalizability of object proposals of the FSHNN based object detector is better than the ablation networks. The results show that though the STDP-STDP performs poorly, it has the least dip in the source-to-target translation ( a drop of 22.5\% in comparison to 24.4\% in Null-STDP and 33\% in BP-Null). This shows that STDP layers help generalize better object proposals

\begin{table}[]
\centering

\caption{Table Describing the Learning Blocks used in the Architectures for \textbf{Ablation Studies}}
\label{tab:layers_abl}
\resizebox{0.5\textwidth}{!}{%
\begin{tabular}{|c|c|c|c|c|c|}
\hline
 &  & \textbf{\texttt{Block 1}} & \textbf{\texttt{Block 2}} & \textbf{\begin{tabular}[c]{@{}c@{}}ResNet\\ Block\end{tabular}} & \textbf{\begin{tabular}[c]{@{}c@{}}FPN\\  Block\end{tabular}} \\ \hline
   \multirow{2}{*}{\textbf{\begin{tabular}[c]{@{}c@{}}STDP-STDP \\ Spiking Obj Det\end{tabular}}} & Layer & SNN & SNN & SNN & SNN \\ \cline{2-6} 
 & Learning & STDP & STDP & BP & BP \\ \hline
  \multirow{2}{*}{\textbf{\begin{tabular}[c]{@{}c@{}}Null-STDP \\ Spiking Obj Det \end{tabular}}} & Layer & None & SNN & SNN & SNN \\ \cline{2-6} 
 & Learning & N/A & STDP & BP & BP \\ \hline
  \multirow{2}{*}{\textbf{\begin{tabular}[c]{@{}c@{}}BP-Null \\ Spiking Obj Det  \end{tabular}}} & Layer & SNN & None & SNN & SNN \\ \cline{2-6} 
 & Learning & BP & N/A & BP & BP \\ \hline
 \multirow{2}{*}{\textbf{\begin{tabular}[c]{@{}c@{}}FSHNN \\ Obj Det\end{tabular}}} & Layer & SNN & SNN & SNN & SNN \\ \cline{2-6} 
 & Learning & BP & STDP & BP & BP \\ \hline
\end{tabular}%
}
\end{table}

\begin{table*}[]
\centering

\caption{Table showing Performance Comparison of the Architectures in \textbf{Ablation Studies}}
\label{tab:perf_abl}
\begin{tabular}{|c|c|c|c|c|c|c|c|c|}
\hline
\textbf{\begin{tabular}[c]{@{}c@{}}Number of \\ STDP Layers\end{tabular}} &  \textbf{mAP} & $\textbf{AP}_{S}$ & $\textbf{AP}_{M}$ & $\textbf{AP}_{L}$ & \textbf{mAR} & $\textbf{AR}_S$ & $\textbf{AR}_M$ & $\textbf{AR}_L$  \\ \hline 
\textbf{6 layer STDP FSHNN}                                                                                                                    & 0.426   & 0.229 & 0.452 & 0.533  & 0.541    & 0.327 & 0.612 & 0.668  \\ \hline
\textbf{\begin{tabular}[c]{@{}c@{}}STDP-STDP \\ Spiking Obj Det \end{tabular}}                                                                                                                     & 0.268   & 0.119 & 0.284 & 0.356  & 0.376    & 0.195 & 0.396 & 0.489  \\ \hline
\textbf{\begin{tabular}[c]{@{}c@{}}Null-STDP \\ Spiking Obj Det \end{tabular}}                                                                                                                     & 0.251   & 0.106 & 0.279 & 0.348  & 0.369    & 0.190 & 0.387 & 0. 481 \\ \hline
\textbf{\begin{tabular}[c]{@{}c@{}}BP-Null \\ Spiking Obj Det \end{tabular}}   &  0.315  & 0.155 & 0.342 & 0.416  & 0.423    & 0.215 & 0.480 & 0.512  \\ \hline
\end{tabular}
\end{table*}

\begin{table}[]
\centering
\caption{AR@100  corresponding  to  different ablation models  trained on  COCO-source-train  and  evaluated  on  different  test  splits.}

\begin{tabular}{|c|c|c|}
\hline
                                                                                 & \textbf{COCO-source-test} & \textbf{COCO-target-test} \\ \hline
\textbf{\begin{tabular}[c]{@{}c@{}}STDP-STDP\\ Spiking Obj Det\end{tabular}}     & 42.5                      & 32.9                      \\ \hline
\textbf{\begin{tabular}[c]{@{}c@{}}Null-STDP\\ Spiking Obj Det\end{tabular}}     & 41.8                      & 31.6                      \\ \hline
\textbf{\begin{tabular}[c]{@{}c@{}}BP-Null\\ Spiking Obj Det\end{tabular}}       & 50.5                      & 33.5                      \\ \hline
\textbf{\begin{tabular}[c]{@{}c@{}}FSHNN Obj Det\\ (6 STDP Layers)\end{tabular}} & 56.2                      & 48.4                      \\ \hline
\end{tabular}

\label{tab:gen3}
\end{table}

%
      

\subsection{Energy Efficiency Comparison}
We next compare the energy advantage of spiking models (FSHNN and Backpropagated spiking layer) with RetinaNet (Table \ref{tab:energy}). As discussed by Kim et al. \cite{kim2020spiking}, most operations in DNNs occur in convolutional layers where the multiply-accumulate (MAC)  operations are primarily responsible during execution. However, SNNs perform accumulate (AC) operations because spike events are binary operations whose input is integrated into a membrane potential only when spikes are received. Thus, SNNs save computational energy due to the use of AC instead of operations in SNNs. 
The number of parameters vs MACs/ACs for the different architectures are summarized in Table \ref{tab:energy}. 

Horowitz et. al.  \cite{horowitz20141} showed that in an in 45nm 0.9V chip, a 32-bit floating-point (FL) MAC operation consumes 4.6 pJ and 0.9 pJ for an AC operation. 
 Based on these measures, we calculated the energy consumption of the spiking object detectors by multiplying  FLOPS  (floating-point operations)and the energy consumption of MAC and AC operations calculated. 
A simplified estimate of the computational energy $E$ for ANN/SNN considering FLOPS count across all $N$ layers of a network can be obtained as \cite{panda2020toward}:
\begin{equation}
\begin{array}{l}
E_{A N N}=\left(\sum_{i=1}^{N} F L O P S_{A N N}\right) * E_{M A C} \\
E_{S N N}=\left(\sum_{i=1}^{N} F L O P S_{S N N}\right) * E_{A C} * T
\end{array}
\end{equation}
For SNN, the energy calculation considers the latency incurred as the rate-coded input spike train has to be presented over $T$ time-steps to yield the final prediction result. Since we are using a training methodology using STDB \cite{rathi2020enabling} which takes much fewer timesteps, we have considered the number of time steps $T=300$ .

\par It should be noted that the prior analysis only considers computational energy savings. More detailed analysis is necessary to understand the total energy advantage of SNN over ANN considering various hardware design issues. For example, 
neuromorphic hardware that uses event-driven computing consumes only static power in the absence of spikes. On the other hand, clock-driven designs benefit much less from sparsity. Also, even on an event-driven neuromorphic chip, there is an additional cost when using spiking neurons - since the neurons must maintain state values over time unlike DNNs which are state-less, their potential needs to be stored, which has an important memory footprint. Further, membrane potential of the neurons must be updated multiple times (once for for every time steps) while evaluating an image thereby incurring additional memory read and write energy dissipations. All of the above factors must be considered for a more accurate analysis of energy advantage of SNN. However, as the focus of this paper is developing the hybrid learning
model for object detection, such detailed energy analysis is beyond
the scope of this paper.

\begin{table}[]
\centering
\caption{Table showing the energy efficiency of object detectors based on SNNs}

\label{tab:energy}
\resizebox{0.85\linewidth}{!}{%
\begin{tabular}{|c|c|c|c|}
\hline
                                                                            & \textbf{Param(M)} &  \textbf{MACs/ ACs}       & \textbf{$EE = \frac{EE_{ANN}}{EE_{SNN}}$} \\ \hline
\textbf{RetinaNet}                                                                   & 52.78             & \textbf{MAC: } $131.41 \times 10^9$ &           N/A                                \\ \hline
\textbf{BPSOD }             & 52.99             & \textbf{AC: }$9.2 \times 10^7$   & 151.53                                    \\ \hline
\textbf{\begin{tabular}[c]{@{}c@{}}FSHNN \\ Obj Det \\ (3 layers STDP)\end{tabular}} & 52.96             & \textbf{AC: }$9.1 \times 10^7$   & 154.88                                    \\ \hline
\textbf{\begin{tabular}[c]{@{}c@{}}FSHNN \\ Obj Det \\ (6 layers STDP)\end{tabular}} & 53.11             & \textbf{AC: } $9.5 \times 10^7$   & 147.54                                    \\ \hline
\end{tabular}%
}
\end{table}



\subsection{\textbf{Comparison to Spiking Yolo}}

We compare FSHNN model performance with the state-of-art spike-based object detector - Spiking YOLO \cite{kim2020spiking}. Though both of them are fully spiking object detectors, there are a few key differences between the two. First, the Spiking Yolo is composed of backpropagated Spiking Convolution Layers that are trained using supervised learning. In contrast, FSHNN couples both STDP based unsupervised learning methodologies with backpropagation learning. This helps the FSHNN network achieve high performance as well as robustness against input noise and less labeled data for training. Second, we analyze the uncertainty and generalizability properties of FSHNN; but such analyses were not performed for spiking YOLO\cite{kim2020spiking}. 
Third, FSHNN shows a higher performance than spiking YOLO.
While the spiking YOLO reported an mAP of $26.24\%$ \cite{kim2020spiking}, the FSHNN network with 6 STDP layers described in this paper achieves an mAP of $42.6\%$. We note that the spiking YOLO object detector is based on the YOLO architecture. The YOLO architecture suffers from an extreme class imbalance where the detectors evaluate a lot of extra candidate locations a majority of which do not contain an object. 
In contrast, FSHNN uses the RetinaNet architecture that fixes this issue by using the Focal Pyramid Network and the Focal Loss.  
The Spiking YOLO is based on the Tiny YOLO object detector architecture which uses $6.97 \times 10^9$ FLOPS of computation. On the other hand, the channel-norm-based spiking YOLO uses $4.9 \times 10^7$ FLOPS of computation in $3500$ time steps. Thus the theoretical energy efficiency achieved using the Spiking YOLO compared to the tiny YOLO is only 1.4x. It must be noted here that the proposed FSHNN based object detectors use almost 4x the number of FLOPS. However, the use of the STDB method greatly reduces the time steps needed from $3500$ to just $300$ as used in this paper. This leads to a huge difference in energy efficiency in the two models.


%% file: tex/conclusions.tex
\section{Conclusion}
We have presented fully spiking hybrid neural network (FSHNN), a novel spiking neural network based object detector by fusing features extracted from both unsupervised STDP based learning and supervised backpropagation based learning. We have also developed used a MC dropout based sampling method to estimate the uncertainty of FSHNN. Experimental results on MSCOCO dataset shows that the FSHNN demonstrates comparable or better performance than standard DNN while promising orders of magnitude improvement in energy-efficiency. Further, we have observed that FSHNN network outperforms the baseline DNN as well as a spiking network trained with backpropagation when tested under input noise or trained with less available labelled data.  
We have also shown that integration of STDP learning helps improve generalization ability of FSHNN. In conclusion, we demonstrated the feasibility of designing a fully spiking network for object detection facilitating the deployment spiking networks for resource-constrained environments. 